\documentclass[notitlepage]{article}
\usepackage{amsmath}
\usepackage{amsfonts}
\usepackage{amssymb}
\usepackage{graphicx}
\usepackage{amsthm}
\usepackage{lineno}%deactivated to try verbatim package 
\usepackage[nottoc]{tocbibind} %p. 48 Mittelbach, adds biblio to toc
\newtheorem*{Capacity--Rate Equivalency Theorem}{Capacity--Rate Equivalency Theorem} 
\newtheorem*{Social Problem Solving Rate Theorem}{Theorem} 
\newtheorem*{Isotropic Network Rate Theorem}{Isotropic Network Rate Theorem} 
\newtheorem*{The Ideal Network Theorem}{The Ideal Network Theorem}
\newtheorem*{General Network Rate Theorem}{General Network Rate Theorem}
\newtheorem*{Capacity Multiplier Theorem}{Capacity Multiplier Theorem}
\newtheorem*{The Natural Logarithm Theorem}{The Natural Logarithm Theorem}
\newtheorem*{The Network Entropy Theorem}{The Network Entropy Theorem}
\newtheorem*{The Network Dating Theorem}{Theorem}
\newtheorem*{Economic Productivity Theorem}{Economic Productivity Theorem}
\newtheorem*{The Basal Rate Network Dating Theorem}{The Basal Rate Network Dating Theorem}
\newtheorem*{Society's IQ Theorem}{Society's IQ Theorem}
\newtheorem*{The Equivalency of the Innate and Bi-nodal Rates Theorem}{The Equivalency of the Innate and Bi-nodal Rates Theorem}
\theoremstyle{definition}
\newtheorem*{The General Collective Problem Solving Capacity Hypothesis}{The General Collective Problem Solving Capacity Hypothesis}\label{GPSCH}
\newtheorem*{General Collective Problem Solving Capacity Problem}{General Collective Problem Solving Capacity Problem}
\newtheorem*{Accelerating Problem Solving Rate Problem}{Accelerating Problem Solving Rate Problem}
\newtheorem*{PSdefinitions}{Problem solving definitions}
\newtheorem*{IQdefinitions}{IQ Definitions}

\begin{document}
\bibliographystyle{plain} %p77, Diller
\pagestyle{myheadings}
\markright{A theory of intelligence}

\title{A theory of intelligence: networked problem solving in animal societies} 
\author{Robert Shour}\
\date{}
\begin{center}
\ {\Large\textbf\ A theory of intelligence: networked problem solving in animal societies} \\
\normalsize
\hfill \\
\ Robert Shour
\hfill \\
\hfill \\
\footnotesize{Toronto, Canada}

\normalsize
\end{center}
\hfill
\begin{abstract}
\noindent
A society's single emergent, increasing intelligence arises partly from the thermodynamic advantages of networking the innate intelligence of different individuals, and partly from the accumulation of solved problems. Economic growth is proportional to the square of the network entropy of a society's population times the network entropy of the number of the society's solved problems. 
\end{abstract}
\


\tableofcontents
\listoftables

\section{Introduction: The General Collective Problem Solving Capacity Hypothesis}\label{Section, Intro}

In this article, I consider the effects of networking on the emergence of intelligence in individuals and societies. The following hypothesis promotes and sustains this investigation: 

\begin{The General Collective Problem Solving Capacity Hypothesis} Society possesses a general, collective problem solving capacity.
\end{The General Collective Problem Solving Capacity Hypothesis}

\textit{The General Collective Problem Solving Capacity Hypothesis} implies that the same general problem solving capacity that society uses, for example, to develop language, is used to solve problems in mathematics, science, business, musical composition and performance, sports contests, social interactions, politics and daily life. ``All life is problem solving'' \cite{KPAll}; all problem solving is a strictly analogous process.

Let's adopt some notational conventions that will allow us to make the observations in the discussion that follows  more precise. The formulas used in the definitions are sometimes modified by a subscript relevant to the context in which they are used.

\begin{PSdefinitions}\label{PSdefn}\hfill
\begin{enumerate}
	\item 	$ \{ PS(t) \} $: a \textit{set} of solved problems, as of time $t$. 
	\item 	$\{PS_{Individual}(t) \}$: the \textit{set} of solved problems that have been learned by an individual, or by a single problem solver in a network of problem solvers, as of time $t$. 
	\item 	$\{PS_{Average}(t) \}$ or $\{PS_{Av}(t) \}$: the \textit{set} of solved problems that have been learned by the average individual in a society, as of time $t$. 
	\item 	$\{PS_{Innate}(t)\}$: a theoretical construct, representing the \textit{set} of solved problems that an individual, who did not have the benefit of any social networking or language, would know, as of time $t$. In past times, such an individual, \textit{with} the benefit of living in a familial group but with only a rudimentary language, would be said to be living in a `state of nature, or that imaginary state, which preceded society' \cite[Book III, Part II, Section II, p. 501]{Hume1739}.
	\item 	$\{PS_{Society}(t) \}$: the \textit{set} of all of a society's---a social network's---solved problems, as of time $t$.
	\item 	$\{PS_{Lang}(t) \}$: the \textit{set} of language problems solved by a society, as of time $t$.
	\item 	$\{PS_{Lex}(t) \}$: the \textit{set} of lexical problems solved by a society, as of time $t$.
	\item 	Lex(t): the \textit{set} of words in a lexicon, as of time $t$. 
	\item 	$|Lex(t)|$: the number of words in a lexicon, as of time $t$.
	\item 	$N(t)$ or $ | \{ PS(t) \}| $: the \textit{number} of solved problems in a set $\{PS(t) \}$, as of time $t$.
	\item 	$|Tot|$: the total number of problems, solved and unsolved, as of time $t$.
	\item 	$r$: a percentage rate of increase per decade in the number of solved problems for some set of solved problems $ \{ PS(t) \} $. For a single period of time, $\Delta t$,
\begin{equation}\label{eq A-10.00}
\begin{split}
\ r &= \frac{d| \{ PS(t) \}|}{dt} \\
&= \left( \frac{| \{PS(t_2)\}| - | \{PS(t_1)\}|}{| \{PS(t_1)\}|}\right) \times 100\%,
\end{split}
\end{equation}
or
\begin{equation}\label{eq A-10.10}
\begin{split}
\ r &= \frac{d N(t) }{dt} \\
&= \left( \frac{ N(t_2) - N(t_1)}{N(t_1)}\right) \times 100\%.
\end{split}
\end{equation}

	\item 	$m$ : a percentage rate such that
	
\begin{equation}\label{Eq InnateRate10.100}
		 m = \frac{d |\{PS_{Innate}(t)\}|}{dt}.
\end{equation}
	 
	\item 	$\Delta t$ from year $t_1$ to year $t_2$ is the number of decades, $(t_2 - t_1)/10$, the number of thousands of years,  $(t_2 - t_1)/1,000$, or the number of years,  as indicated by the context.
\end{enumerate}
\end{PSdefinitions}

The hypothesis raises the 

\begin{General Collective Problem Solving Capacity Problem} Is there any way to prove \textit{The General Collective Problem Solving Capacity Hypothesis}---that is, to show that the same problem solving capacity of a society is used to solve all types of problems?
\end{General Collective Problem  Solving Capacity Problem}

We want to find a \textit{common attribute} of different kinds of collectively solved problems that will demonstrate that society has a general, collective, problem solving capacity. Any such attribute should be \textit{objectively measurable}; otherwise \textit{objective comparison} is not possible. To narrow our scope of inquiry into what the \textit{common attribute} might be, we consider how to characterize a problem solving capacity.

If \textit{Society A} solves twice as many problems in a year as \textit{Society B}, \textit{Society A} has demonstrated, literally, twice the problem solving capacity---solved problem productivity---of \textit{Society B}. Let a subscript $A$ indicate a set of solved problems pertaining to \textit{Society A} and let a subscript $B$ indicate indicate a set of solved problems pertaining to \textit{Society B}. Here, we consider the increase,

\begin{equation}\label{eq A-10.20}
\\ | \{PS(t_2)\}_{A}| - | \{PS(t_1)\}_{A}|=\Delta \left(\left|\left\{PS\right\}_{A}\right|\right)
\end{equation}

\noindent over a period of time, in the existing number of solved problems for a \textit{Society A}, as indicative of \textit{Society A's} problem solving capacity. If the number of a society's solved problems remained unchanging from one year to the next, then that society would not have a problem solving capacity, just an unchanging store of solved problems. 

Suppose, in hypothetical circumstances, that the population of a human society is unvarying---constant---and that the proportion of the population that are problem solvers in the society is constant. Then, even if the number of problem solvers affects a society's solved problem output, in these circumstance the number of problem solvers is not a factor.  If \textit{Society A} has $k$ times the problem solving capacity of \textit{Society B} during the \textit{same period of time}, then, adapting (\ref{eq A-10.20}),

\begin{equation}\label{eq A-10.30}
	 \Delta \left(| \{PS\}_{A}|\right) = k \times \Delta \left(| \{PS\}_{B}|\right).
\end{equation}

The ratio between the left and right sides of (\ref{eq A-10.30}), of the number of problems solved in two different societies during the same period of time, is maintained for the rate at which problems are solved, that is,

\begin{equation}\label{eq A-10.40}
	 \frac{ d | \{PS\}_{A}|}{dt} = k \times \frac{d | \{PS\}_{B}|}{dt}.
\end{equation}

If society has a general problem solving rate, then if its general problem solving rate results in increasing the number of solved problems of one kind, then the number of solved problems of all other kinds should also increase at the same rate. If we can identify different kinds of  general collective problems for which the problem solving rates are the same, then we have evidence in favor of---and consistent with---the existence of society's general collective problem solving capacity. 

In light of the foregoing discussion, we have found one possible \textit{common attribute} that we can use to test our hypothesis that society has a general collective problem solving capacity: the rate of collective problem solving.

\section{The plausibility of a general collective rate of problem solving}\label{Section plausibility}

\subsection{General considerations}

Suppose  the capacity of any individual to solve any \textit{particular} kind of problem is an instance of that individual's general problem solving capacity. Then one can infer that the same applies to society, that is, that \textit{the capacity of a society} to solve a particular kind of collective problem is an instance of society's general collective problem solving capacity. For if \textit{individuals} apply the same \textit{average amount} of energy to solve an average problem, then society collectively will apply the same \textit{average cumulative amount} of energy to solve an average collective problem. 

There is accepted evidence that implies that individuals use the same average amount of energy to solve the average problem---that individuals have a general problem solving capacity---that is characterized by experts in the field of intelligence testing as a general intellectual proficiency. The

\begin{quote}
``tendency shared by most individuals to perform many different intellectual tasks at about the same level of proficiency \ldots has been demonstrated repeatedly in statistical analyses of the interrelationships of performances on tests measuring different intellectual functions \ldots'' \cite[p. 22]{Lezak1983}. 
\end{quote}

What has been demonstrated for individuals necessarily applies with greater force to society. In a society, differences in the problem solving capacities of individuals---for some the capacity is higher, and for others the capacity is lower, than average---offset each other. The larger the sample of problem solvers, the more likely it is that the average amount of energy spent by the average individual problem solver in solving an average problem is a good \textit{estimate} of the society's average amount of energy spent per individual solving the average problem. Then, for collective problems, the cumulative amount of energy required should be the same for different kinds of problems. This is a consequence of \textit{The Law of Large Numbers}\ when applied to \textit{the number of problem solvers}. The larger the sample of problem solvers, the more likely it is that the \textit{estimated} average individual problem solving capacity equals the \textit{actual} average individual problem solving capacity for the entire society: 

\begin{quote}
	``\ldots as the number of observations increases, so the probability increases of obtaining the true ratio between the number of cases in which some event can happen and not happen, such that this probability may eventually exceed any given degree of certainty'' \cite[p. 328]{Bernoulli2006}.
\end{quote}

\noindent In principle, it should be possible to estimate the actual amount of energy used by the average problem solver to solve the average problem, \textit{from the point of view of the problem solvers}.

The second reason why using average problems as data is likely to reveal society's average collective problem solving rate involves the thermodynamics of problem solving.

If it requires a certain amount of energy to solve a problem, then to solve a set of problems requires a certain \textit{total} amount of energy. If we assume that the number of problems solved by a society is proportional to the average amount of energy used to solve each problem, then---considering a solved problem as equivalent to information---

\begin{equation} \label{eq 4-4}
\\	 energy \ input \propto \ information \ output .
\end{equation}

\noindent If we denote the quantity of information for \textit{Subject} \# 1 by $(Information)_{1}$, and the amount of energy used to create that information by $(Energy)_{1}$, and similarly for \textit{Subject} \# 2,  the proportionality of information and energy may be expressed  as

\begin{equation}\label{eq 5}
	 \frac{(Information)_{2}}{(Information)_{1}}=\frac{(Energy)_{2}}{(Energy)_{1}}.
\end{equation}

\noindent We might put it this way: \textit{Sets of solved problems containing the same amount of information required the same amount of energy to solve them.} If

\begin{equation}\label{eq Inf-eq-10.10}
\\ (Information)_{k}=(Information)_{j}
\end{equation}

\noindent for sets of information, then

\begin{equation}\label{eq Inf-eq-10.20}
	(Energy)_{k}=(Energy)_{j},
\end{equation}

\noindent for all $j$ and $k$, where (\ref{eq Inf-eq-10.10}) holds. 

The equivalence of ratios in (\ref{eq 5}) resembles the form of the relationship between heat and absolute temperature in thermodynamics,

\begin{equation}\label{eq 6}
\frac{Q_{2}}{Q_{1}}=\frac{T_{2}}{T_{1}} ,
\end{equation}

\noindent where, in an ideal heat engine, 
\begin{itemize}
	\item $Q_{2}$ represents an amount of heat from a heat source  added to the heat engine's working substance, 
	\item $T_{2}$ represents the absolute (higher) temperature of that heat source, 
	\item $Q_{1}$ represents the amount of heat removed from the heat engine's working substance and added to the engine's heat sink, and 
	\item $T_{1}$ represents the (lower) absolute temperature of the heat sink.
\end{itemize}
\noindent(A good account of the ideal heat engine is in \cite[Ch. 44, Vol. 1]{Feynman1963}.)

If the average problem solver has an average general problem solving capacity, then it must be that the same energy input can achieve, regardless of the nature of the problem, the same quantity of information output. That would imply that the analogy of the equality in (\ref{eq 5}) to the equality in (\ref{eq 6}) is exact. Whether society has an average general collective problem solving capacity is equivalent to asking whether  (\ref{eq 5}) is true for \textit{different} kinds of information obtained by general \textit{collective} problem solving. 

 If a society as a whole confronts a particular problem, the society is more efficient if it diverts its problem solving energy resources to solving that problem only until the payoff---the benefit of obtaining a solved problem---for the energy required to solve the problem matches the payoff for alternative uses of society's problem solving energy resources. If a society is adaptive, the competition among problems for the society's finite problem solving energy resources should result in the same average level of solved problem productivity for \textit{different kinds} of collective problems that have the same information output. 
 
New solutions of problems, and improvements in existing solutions, are \textit{inventions}---conceptual inventions---by problem solvers. Problems that confront society compete for society's problem solving energy resources. Kenneth Arrow remarked that

\begin{quotation}
	``Invention is here interpreted broadly as the production of knowledge. From the viewpoint of welfare economics, the determination of optimal resource allocation for invention will depend on the technological characteristics of the invention process and the nature of the market for knowledge'' \cite{Arrow1959}.
\end{quotation}

\noindent Society collectively determines the `optimal resource allocation' for the solving of its \textit{problems}---how to allocate its resources to create knowledge. 

 The efficiency criterion for the allocation of problem solving resources implies that problem solving outputs for different kinds of problems should be, on average, proportional to their energy inputs, as in (\ref{eq 5}). In principle, there \textit{should} exist  an average amount of energy per average general collective problem, from \textit{the point of view of the collective problems}.

Since the number of solved problems is finite and in principle, enumerable, we can number the solved problems sequentially, using 1, 2, 3, \ldots, N$(t)$; the subscript \textit{i} pertains to a problem's assigned number. Each solved problem, $(PS)_i$ required a finite number $\xi_i$ of energy units $\epsilon$ to solve it; a standard energy unit, $\epsilon$, is selected so that when we add up the total amount of energy required to solve all $N(t)$ solved problems, the total amount of energy is

\begin{equation}\label{eq 7}
\sum^{N(t)}_{i=1} \xi_{i}\epsilon =N(t) \epsilon.
\end{equation}

\noindent The average energy per solved problem is

\begin{equation}\label{eq 8}
\\\frac{\sum^{N(t_1)}_{i=1} \xi_{i}\epsilon}{N(t_1)}=\frac{N(t_1)\epsilon}{N(t_1)}=\epsilon.
\end{equation}

Therefore, if there are a finite number of solved problems which each required a finite amount of energy to solve, then it must be possible in principle to calculate the average amount of energy that was required to solve the average problem. For there to be the possibility of calculating an average, it is necessary that the total number of solved problems, and the total amount of energy required to solve them, \textit{both} be finite. A human society and its store of solved problems satisfy those requirements. 

Human beings, for example, consume a finite amount of food, which supplies a finite amount of energy, during their lifetime. If (1) the average person consumes 2000 calories per day, (2) the proportion of daily calorie intake devoted to problem solving is known, and (3) if the number of problems solved over a period of time is known, \textit{then} the average amount of energy used to solve the average problem could in principle be calculated. Similarly, the average amount of energy used to transmit the average amount of information from one individual directly to another could, in principle, be calculated. 

All of these considerations imply that it is plausible that a society has a general collective problem solving rate.

\subsection{Analogical generalizations}

A model that successfully describes general collective problem solving for human societies should apply as well to  general collective problem solving by other animal societies; general collective problem solving by human societies is a paradigm for  general collective problem solving for all societies of animals. Studying collective problem solving in human societies has advantages over studying problem solving in other animal societies.  Statistics about changes in problem solving capacities and economic productivity are more numerous and more available for societies of human beings than for societies of other animals.

We can study the emergence of general collective problem solving in human societies,
\begin{itemize}
	\item  as an analogy for emergent networked processes generally,
	\item  as a paradigm for networked problem solving in animal societies, and
	\item  as a phenomenon consistent with the concepts and theories of thermodynamics.
\end{itemize}

\section{Estimating a problem solving rate}\label{Section, estimating}

\subsection{Rate estimation}\label{subsec, rate est}

In this section, we consider some aspects of finding a problem solving rate. We assume that in principle it is possible to enumerate all solved problems that an individual has learned, or that a society has accumulated in a store of solved problems, leaving aside the difficult problem of what counts as a single solved problem. 

The output of a problem solving capacity consists of solved problems. When society successfully applies its problem solving capacity to solve a problem, society increases the number of \textit{already} solved problems. If the number of solved problems increases \textit{linearly}, then for any number of decades in the period of time, $\Delta t$, 

\begin{equation}\label{eq A-70}
\begin{split}
	 \left|\{PS(t_2)\}\right| &= \left|\{PS(t_1)\}\right|+\left( \frac{ d \left|\left\{PS(t_1)\right\}\right|}{dt} \times \Delta t \times  \left|\{PS(t_1)\}\right|\right) \\
&= \left(1+ \left[ \frac{ d \left|\left\{PS(t_1)\right\}\right| }{dt} \times \Delta t\right] \right)  \times \left|\{PS(t_1)\}\right| .
\end{split}
\end{equation}

If the number of problems increases \textit{exponentially}, then for any period of time $\Delta t$, 

\begin{equation}\label{eq A-80}
\\ \left|\{PS(t_2)\}\right| =  \left(1+ \frac{ d \left|\left\{PS(t_1)\right\}\right|}{dt}\right) ^{ \Delta t} \times \left|\{PS(t_1)\}\right|,
\end{equation}

\noindent or, having regard for the definition of $r$ in (\ref{eq A-10.10}),

\begin{equation} \label{eq A-80.10}
\\ \left|\{PS(t_2)\}\right| = (1+ r)^{ \Delta t} \times \left|\{PS(t_1)\}\right|.
\end{equation}

Like interest accruing on principal, the number of newly solved problems increases the total number of solved problems exponentially. If the accumulation of solved problems is analogous to the accumulation of interest, then it is likelier that (\ref{eq A-80}) and (\ref{eq A-80.10}) rather than (\ref{eq A-70}) characterize the way that solved problems accumulate. 

If we know the number of solved problems at an earlier year $t_1$ and also at a later year $t_2$, it is possible to calculate the \textit{average} rate $r$ per decade, of exponential growth in the number of society's solved problems, by solving for $r$ in

\begin{equation}\label{eq 9}
\\ N(t_2)=N(t_1) \exp(r \cdot \Delta t). 
\end{equation}

\noindent Then, from  (\ref{eq 9}), with $\left|\{PS(t_1)\}\right| = N(t_1)$ and $\left|\{PS(t_2)\}\right| = N(t_2)$, 

\begin{equation}\label{eq 10}
\\r=\left( \ln \left[\frac{N(t_2)}{N(t_1)}\right] \right)\  \div\Delta t .
\end{equation}

\noindent (\ref{eq 10}) suggests three kinds of data can be used to determine society's average rate per decade of collective problem solving $r$: 

\begin{itemize}
	\item Data that has measured $r$ itself, or a related rate equal to $r$;
	\item Data that has measured $N(t_1)$ and $N(t_2)$ for some set of solved problems, which allows us to find the logarithm of their ratio as in (\ref{eq 10}), and to solve for $r$; or 
	\item Data that gives the ratio 

\begin{equation}\label{eq 11}
	 \frac{N(t_2)}{N(t_1)}
\end{equation}

\noindent but not the values of $N(t_1)$ and $N(t_2)$ separately, which allows us to solve for $r$ in (\ref{eq 10}). 

\end{itemize}

With the rate per decade $r$ determined directly as in (\ref{eq 10}) when the values of $N(t_1)$ and $N(t_2)$ are known, or indirectly when only the ratio set out in (\ref{eq 11}) is known, we can calculate the average rate of increase $r$ in the number of solved problems per decade. If, using data for different kinds of solved problems, we calculate the same rate $r$ for them, we will have obtained evidence favoring the validity of the hypothesis that society has a general collective problem solving rate. The closer that the values of the rates for the different kinds of solved problems are to each other, the more persuasive such evidence is.

This is a refinement of the possible common attribute, the rate of general collective problem solving, for testing \textit{The General Collective Problem Solving Capacity Hypothesis}. We can test not just whether the rates coincide, but we can impose a tougher test of whether \textit{exponential rates} coincide over an extended period of time. If rates for two different kinds of problem solving are both exponential, but the functions describing each differ, the gap between the two rates would likely increase over time. If the functions coincide over a long period of time, the hypothesis has passed a sterner test.

As a consequence of \textit{The Law of Large Numbers}, and from  (\ref{eq 11}), it appears that when $N(t_1)$ and $N(t_2)$ are large, we will have more confidence that when their ratio in (\ref{eq 11}) is used to solve for $r$ in (\ref{eq 10}), we will obtain an estimate $r_{Est}$ that is close to society's actual  average general problem solving rate, $r_{Act}$. 

\subsection{Error estimation}\label{subsec, error est}

Suppose that we have estimates of both $ N(t_1)$ and $N(t_2)$, and that we use the equation in (\ref{eq 10}) to estimate $r$. How accurate is $r_{Est}$ as an estimate of the rate $r$?

By assuming the existence of an actual rate, $r_{Act}$, we can show that the longer the time period $\Delta t$ used for estimating that actual average rate, the closer $r_{Est}$ is to $r_{Act}$, all other things being equal. 

Suppose $r_{Act}$ and $\Delta t=(t_2-t_1)$ are given, and that $\left[N(t_{1} )\right]_{Act}$ is the actual value of $N(t_{1} )$. Let 

\begin{equation}\label{eq 12}
\\ \left[N(t_2)\right]_{Act}
\end{equation}

\noindent be the actual number of solved problems at year $t_2$. Let $r_{Est}$ be an estimate of  society's problem solving rate, and let $\left[N(t_2)\right]_{Est}$ be an estimate of the number of solved problems at year $t_2$. If we use an estimated average rate, $r_{Est}$, to estimate $\left[N(t_{2} )\right]_{Act}$ based on $\left[N(t_{1} )\right]_{Act}$, then, applying (\ref{eq 9}),

\begin{equation}\label{eq 13}
\\\left[N(t_2)\right]_{Est}=\left[N(t_1) \right]_{Act} \exp(r_{Est}\cdot \Delta t).
\end{equation}

\noindent Let 

\begin{equation}\label{eq 14}
	 \Delta N_2 = \left[N(t_2)\right]_{Est}-\left[N(t_2)\right]_{Act}.
\end{equation}

\noindent Here $\Delta N_2$ is the error in the estimate of the size of $\left[N(t_2)\right]_{Act}$ resulting from using an inaccurate $r_{Est}$ applied to $\left[N(t_1)\right]_{Act}$ to obtain a value for $\left[N(t_2)\right]_{Est}$ as an estimate of $\left[N(t_2)\right]_{Act}$. Taking the ratio $\left[N(t_2)\right]_{Est}$ to $\left[N(t_2)\right]_{Act}$ reveals that a longer time period reduces the amount of error in $r_{Est}$. For, applying (\ref{eq 13}), 
\begin{equation}\label{eq 15}
\begin{split}
	 \frac{\left[N(t_2)\right]_{Est}}{\left[N(t_2)\right]_{Act}} 
& =\frac{\left[N(t_2)\right]_{Act}+ \Delta N_2}{\left[N(t_2)\right]_{Act}} \\
&=\frac{\left[N(t_1) \right]_{Act} \exp(r_{Est}\cdot \Delta t)}{\left[N(t_1) \right]_{Act} \exp(r_{Act}\cdot \Delta t)} \\
&=\frac{\exp(r_{Est}\cdot \Delta t)}{\exp(r_{Act}\cdot \Delta t)}\\
&=\exp((r_{Est}-r_{Act}) \cdot \Delta t).
\end{split}
\end{equation}

\noindent Taking, in  (\ref{eq 15}),  the logarithm of the right side of the first line and the logarithm of the last line, we find that 

\begin{equation}\label{eq 16}
\\\ln \left(\frac{\left[N(t_2)\right]_{Act}+ \Delta N_2}{\left[N(t_2)\right]_{Act}}\right)=(r_{Est}-r_{Act}) \cdot \Delta t .
\end{equation}

From (\ref{eq 16}), the error in the estimate of $r_{Act}$ resulting from using $r_{Est}$ to estimate $\left[N(t_2)\right]_{Act}$ is

\begin{equation}\label{eq 17}
\\ r_{Est}-r_{Act}=\left[ \ln \left(1+\frac{ \Delta N_2}{\left[N(t_2)\right]_{Act}}\right) \right] \div \Delta t .
\end{equation}

\noindent The error in an estimate of $r_{Act}$ becomes smaller as  $\Delta t$ increases (and also as $\Delta N_2$ decreases). In percentage terms, the error is

\begin{equation}\label{eq 18}
\\\frac{r_{Est}-r_{Act}}{r_{Act}} \times 100\%.
\end{equation}

As an example of the effect of the number of decades on an error in estimating $r_{Act}$, suppose the actual number of solved problems $\left[N(t_2)\right]_{Act}$ is known for a year $t_2$ which is 33.2 decades later than $t_1$. If $\left[N(t_2)\right]_{Est}$ is 10\% higher than $\left[N(t_2)\right]_{Act}$ as a result of imperfect data---$\Delta N_{2}$ is 10\% of $\left[N(t_2)\right]_{Act}$---an actual rate of problem solving of .0341 (3.41\%), for example, per decade will be estimated to be higher by about 

\begin{equation}\label{eq 19}
\begin{split}
	 r_{Est}-r_{Act}
&=\left[ \ln \left(\frac{(1.1)\left[N(t_2)\right]_{Act}}{\left[N(t_2)\right]_{Act}}\right) \right]\div \Delta t \\
&= \left[\ln (1.1)\right]/33.2 \\
&=0.002871 .
\end{split}
\end{equation}

\noindent The error $0.002871$ as a proportion of $r_{Act}$ , which we here suppose to be 3.41\% per decade, would therefore be 

\begin{equation}\label{eq 20}
\\0.002871\div .0341\ =.08418,
\end{equation}

\noindent or about 8.4 percent. If $\left[N(t_2)\right]_{Est}$ =$1.1 \times \left[N(t_2)\right]_{Act}$---if $\left[N(t_2)\right]_{Est}$ is 10\% too high an estimate---over a period of 332 years, the \textit{estimated} average rate per decade, $r_{Est}$, would be $3.41\% + 0.2871\% = 3.6971\%$ per decade, about 8.4\% more than the actual value of 3.41\% per decade used in this example. If the span of time was 839 years---83.9 decades---then a 10\% error in $\left[N(t_2)\right]_{Est}$ would result in an error in the estimate of $r_{Act}$ of about 3.33\%. If the span of time was 3,742 years---374.2 decades---then a 10\% error in $\left[N(t_2)\right]_{Est}$ would result in an error in the estimate of $r_{Act}$ of .007469, which is less than three quarters of one percent. The results of error calculations when $\left[N(t_2)\right]_{Est}$ is 10\% too high are summarized in Table \ref{Table 1-1}.

\renewcommand{\arraystretch}{1.25}
\begin{table}[ht]
\begin{center}
	\begin{tabular}{|c|c|}\hline
	Number of years & Size of error in \% \\\hline
	100 & 27.9 \\\hline
	332 & 8.418 \\\hline
	839 & 3.33 \\\hline
	3742 & 0.7469 \\\hline
	\end{tabular}
	\caption{Percentage error in $r_{Est}$ due to $\left[N(t_2)\right]_{Est}$ being 10\% too high.}\label{Table 1-1}
	\end{center}
\end{table}

If $\left[N(t_2)\right]_{Est} = 0.9 \times \left[N(t_2)\right]_{Act}$---if $\left[N(t_2)\right]_{Est}$ is 10\% too \textit{low} an estimate---over a period of years, we can do a similar set of calculations leading to the results in Table \ref{Table 1-1Lower}.

\renewcommand{\arraystretch}{1.25}
\begin{table}[ht]
\begin{center}
	\begin{tabular}{|c|c|}\hline
	Number of years & Size of error in \% \\\hline
	100 & 30.9 \\\hline
	332 & 9.3 \\\hline
	839 & 3.6 \\\hline
	3742 & 0.826 \\\hline
	\end{tabular}
	\caption{Percentage error in $r_{Est}$ due to $\left[N(t_2)\right]_{Est}$ being 10\% too low.}\label{Table 1-1Lower}
	\end{center}
\end{table}

If the error in the number of problems at year $t_2$ is \textit{less than 10\%}, whether high or low, then the size of the errors in Tables \ref{Table 1-1} and \ref{Table 1-1Lower} would be even less. This discussion suggests that, \textit{all other factors being equal}, a rate estimated by using a longer period of time is more reliable. Similar calculations apply to determining the effect of an error in estimating  $\left[N(t_1)\right]_{Act}$---the actual number of solved problems at year $t_1$. 

This error analysis suggests preferring data consisting of a large number of problems evaluated by a large number of people over a long period of time, to improve the accuracy of our estimate of society's average general collective problem solving rate.

\section{Testing the general collective problem solving capacity hypothesis: required attributes of data}\label{Section, Testing}

\subsection{Kinds of data}\label{Subsection Kinds of Data}

One possible way of estimating society's average, general, collective problem solving capacity---society's average, general, collective problem solving rate---is to identify a kind of solved problem that can be counted, and to then calculate the problem solving rate as described in subsection \ref{subsec, rate est}. 

For a collection of solved problems to qualify as general collectively solved problems they must be intrinsically an accomplishment of society acting as a single problem solving entity. The total number, per year, of books published, words printed, things invented or manufactured, might be representative of the problem solving output capacity of a society. A language is a good example of a set of solved problems that involves all of society. The development of language is intrinsically an accomplishment of a society. A language is a collective, emergent phenomenon. It is, in its essence, a network phenomenon---a medium for transmitting information within a network. To use language as a way of testing society's general collective problem solving productivity requires us to identify a feature of language that consists of a countable subset of solved language problems, with data available for different points of time in order to estimate the problem solving rate. 

By comparing the cumulative number of solved problems for one kind of solved problem as of two different years, we convert the problem of measuring the implicit energy content of a set of solved problems to one of \textit{finding a countable feature} of a particular set of solved problems. What counts as a separately solved problem? We need to find a set of solved problems where the criteria for deciding what counts as a solved problem are relatively consistent and easy to determine. It would be most helpful if we could identify a set of already compiled \textit{and counted} solved problems. 

Suppose  it is possible to enumerate a set of solved problems. Suppose that, at an earlier year $t_1$, a society has an accumulated store of $N(t_1)$ solved problems, where

\begin{equation}\label{eq 6b}
	 N(t_1)= |\{PS_{Society}(t_1)\}|.
\end{equation}

To test the statement implied by (\ref{eq 5})---the amount of information is proportional to the amount of energy used to create that information---we need to numerically compare the implicit energy contents of different bodies of information. To compare their implicit energy contents, we need to assign a number to their respective amounts of information. Since it is difficult to compare \textit{different kinds} of problem solving, such as lexical problem solving and mathematical problem solving, by means of a common criterion, we seek statistics that correspond to the amount of information for \textit{the same kind of problem solving} at different points in time. 

Another possible way of estimating society's average, general, collective problem solving rate might be to use a statistic that is, \textit{indirectly}, equivalent to it. We will explore that possibility in this section as well. Examples of statistics about averages that might be useful are the average test scores of individuals in IQ testing, and the average economic productivity of individuals in a society.

We will gain confidence in the correctness of our hypothesis that society has a general collective problem solving capacity if we find similar general collective problem solving rates for different kinds of general collective problems. The closer the rates are to each other, and the longer the period of time during which they are close to each other, the more confidence we will have. We therefore seek representative statistics to measure society's average general collective problem solving rate. In this section, we will consider the criteria for choosing such data. First we consider the evaluative nature of society's general, collective---and enumerable---problem solving rate. Then we consider how it is possible to use statistics about an \textit{average individual} problem solving rate for comparison to society's general \textit{collective} problem solving rate.

\subsection{Economic growth data}\label{Subsection Ec data}

Economic statistics are a possible source of evidence about society's general collective problem solving rate. If society's store of solved problems increases, then economic growth should also increase. If a society's general collective problem solving capacity results in an increase of the number of its solved problems, that should improve the economic well-being of its citizens. We might put it this way:

\begin{center}
\footnotesize
$Applying \ a \ problem \ solving \ capacity  \Rightarrow the \ number \ of \ solved \ problems\ increases, $
\end{center}

\noindent and

\begin{center}
\footnotesize
 $the \ number \ of \ solved \ problems\ increases \Rightarrow the \ economy \ grows$.
\end{center}

\normalsize

This chain of inference suggests that \textit{individual intelligence} and economic growth are related, since society's general collective problem solving capacity is the networked \textit{general} problem solving capacity of its \textit{individual} citizens. 

Individual intelligence---both learned and innovative problem solving---is used to develop and to evaluate a society's improved and new technologies. Economists infer that improvements in technology enable economic growth \cite{Romer1990}. Since improvements in technology derive from the application of the problem solving capacities of individuals and of society to increase the productivity of some process, then so do improvements in the  economy. Increasing society's intelligence---its general collective problem solving capacity---should increase economic growth. If we can improve our understanding of intelligence, then we may improve our understanding of how society's general collective problem solving capacity leads to economic growth. 

Intelligence and economic growth are also related in another way: both are the product of emergent processes. A brain's networked neurons manifest themselves as \textit{one single mind}. Similarly, society's---a social network's---
\begin{quote}
``dispersed bits of incomplete and frequently contradictory knowledge which all separate individuals possess'' \cite[p. 77]{Hayek1948}
\end{quote}

\noindent leads, in an economy, to a market solution that

\begin{quote}
``might have been arrived at by one single mind possessing all the information'' \cite[p. 86]{Hayek1948}.
\end{quote}

Individual intelligence and the development of a market economy are both emergent processes. In emergence, disparate but networked parts arrange themselves to have a capacity that none of the parts individually has. If we can improve our understanding of how networking plays a role in the emergence of \textit{individual} intelligence, we may improve our understanding, by analogy, of how markets and \textit{collective} intelligence emerge. So, for economists, there are at least two reasons to study the nature of intelligence: first, to relate intelligence to economic growth, and second, because how individual and collective intelligence emerge may provide an analog for how markets emerge.

\subsection{The evaluative nature of collective problem solving}\label{Subsection PS as evaluation}

If we decide to use enumeration as the way to estimate $r$, then the type of solved problem that is the object of our investigation must have a definite solution that is discrete and countable. Problems concerned with political and social interactions often do not qualify as discrete and countable. In contrast, problems in a mathematics test, or inventions invented during a period of time, qualify as being discrete, countable, and capable of being identified as being solved. 

Society, mostly, does not \textit{collectively} invent new technologies like new light bulb designs. There are too few technological problems confronted by too few individuals who, as inventors, have problem solving capacities that are too unrepresentative of the average person, for us to consider most technological innovations and inventions as indicative  of society's average general collective problem solving rate. It must be the case that society's average  general collective problem solving rate measures the rate at which members of society solve evaluative problems. Even for language, which is a communication technology for a social network, created through the solved problem contributions of many individuals, evaluation plays a prominent role.

For a word, society's evaluative consensus is required before it becomes part of the lexicon. A new word presents word users with these problems: is this new word useful, convenient, efficient, and an improvement on the available alternatives, so that it is worth the effort required to learn it? For a consumer product, the consumer's problem is not, how can I invent this? The consumer's problems are instead: is this more useful, convenient and efficient than available alternatives? should I buy this?  For a new scientific theory, a scientist's problems include: should I accept this new theory and adjust my research, texts, and teaching accordingly \cite{Kuhn1970}. Legislators must solve evaluative problems such as, does this proposed legislation remediate a social, economic, or political issue without itself causing harm? if passed, will this legislation be viewed favorably by my constituency? Jurists must solve problems that include: which solution is consistent with existing legal principles? which possible solution best resolves the legal issue involving the parties? which resolution is consistent with an appropriate remedy? Society must also evaluate how to  best receive, store, retrieve, process, transmit and record solved problems. We infer that society's general collective problem solving is evaluative. If a varied or new idea does not pass society's evaluative muster, for most practical purposes, it has no effect. If no one pays attention to an idea, the idea might as well not exist. 

Consider encoding a computer software program into computer code as analogous to encoding abstractions into words.  Computer software continually improves as the result of the efforts of thousands of computer software engineers. Language, analogously, also continually improves, but emergently, and over much longer periods of time involving all of a society; language is a large-scale, open source,\footnote{`Open source' is an observation of Michael Shour, Toronto.} software engineering project. 

Just as energy is spent to devise computer software, so also is energy spent to improve language. 

For an ensemble of independent systems which are uniformly random, James Clerk Maxwell, Ludwig Boltzmann and J. Willard Gibbs showed in the late 1800s that the statistically most likely state is one in which the same average amount of energy occupies equal volumes. By analogy, the same amount of energy being required to solve the average problem is a statistically more likely state. Considering the distribution of energy among solved problems, which each required the same average amount of energy to solve, is analogous to considering the distribution of energy among equally sized `volume elements'  \cite[p. 50]{BoltzGs} or `region elements' \cite[p. 124]{Planck1914} in statistical mechanics. 

To demonstrate that the same amount of energy is required to solve \textit{any kind} of \textit{average} problem, as distinguished from specialized non-collective problems, we need to use, as data, evaluative problems that any person in society can solve, and which are evaluated by many people. It should not matter what the nature of the problem is, or who the particular problem solver is. Specialized problem solving will not reveal society's \textit{average general collective} problem solving rate. If we successfully choose as data average collectively solved problems, there is an increased likelihood that society's average general collective problem solving rate will be revealed. This increased likelihood arises for the reasons set out in Section \ref{Section plausibility}.

Society evaluating the merits of a proposed solution to a given collective problem is equivalent to networking the individual evaluations of all members of society. If a large number of people contemporaneously evaluate the merit of a proposed solution to a given problem, and if each person solves  several sub-problems in such an evaluation, then $N(t)$, the total number of such solved problems, is large.  Just as, in recording the results of a large number of flips of a coin, we expect the ratio of heads to tails to better reflect the true proportion of the chance of heads compared to the chance of tails, so we expect that a large number of solved evaluative problems produced by a large number of people will better enable us to \textit{estimate} society's \textit{actual} average general collective problem solving rate. 

Therefore, measurement of society's average general rate to solve  \textit{evaluative} problems can help test the hypothesis that a society has an average general, collective, problem solving capacity. 

\subsection{When an average of individual rates coincides with a collective rate}\label{Subsection PS-av=PS-soc}

\subsubsection{Problem solving in IQ tests}\label{Subsubsection about IQ tests}

Statistics about average IQ test scores are available. IQ tests are normalized by the producers of IQ tests---adjusted so that the average IQ of a contemporaneous reference group is at all times 100. The rate of change in the normalization has been measured. Therefore, it is possible to measure the rate at which average IQs increase by comparing normalizations of average IQ test results for IQ tests administered at earlier times to average IQ test results for IQ tests administered at later times. We look at IQ tests because they are a particular instance of indirectly measuring an average individual general problem solving rate. Since some statistics about average IQs are available, the rate at which average IQs increase can be compared to society's average general, collective problem solving rate in other areas, such as lexical growth and lighting efficiency. 

An IQ test has a standardized set of test questions, designed to measure an individual's problem solving capacity for different kinds of problems, to be completed over a prescribed amount of time. The IQ test has two fixed variables---the test questions and the amount of time to complete them---out of three. The third variable, the number of correctly solved problems $| \{PS\} |$ out of the total number of problems $|Tot|$, helps determine the IQ test scores of each person because both sets, $\{PS\}$ and the set of problems included in $|Tot|$, are enumerable---can be counted. The proportion,

\begin{equation}\label{eq A-60}
\\ \frac{|\{PS\}|}{|Tot|},
\end{equation}

\noindent can be measured and compared to a contemporaneous standardized average proportion of problems correctly solved by a reference group of people who have taken the test. The test, having sampled an individual's problem solving capacity by questions designed to test different kinds of problem solving skills, indirectly and in part, estimates (we infer), for the person tested, the proportion of society's knowledge---the proportion of solved problems---that the person has learned. 

An individual's problem solving capacity is applied to, among other kinds of problems, the problem of how to learn society's solved problems---how to acquire information from other individuals by socially networking with them, and from society's store of solved problems. If, consistent with \textit{The General Collective Problem Solving Capacity Hypothesis}, an \textit{individual's} problem solving capacity is a general capacity, it follows that the rate at which an individual learns society's solved problems is approximately equal to the person's general problem solving rate. Therefore, measuring what a person has learned is an indirect way of measuring an individual's general problem solving rate. If a person's problem solving rate is positive, then, if it is consistently applied, the person's own store of  solved problems, acquired to a large extent by learning society's solved problems, should increase. 

If the intelligence of a person is equivalent to the person's general problem solving capacity, then if an IQ test estimates the person's general intelligence, it at the same time measures the person's general problem solving capacity. If an IQ test is an accurate estimate of a person's intelligence, then informally we might speak of `a person's IQ' as if we are speaking of that person's intelligence. But an IQ test, even in this informal sense, is only \textit{an estimate} of the person's `actual IQ.'

An IQ test result, as expressed in (\ref{eq A-60}), is analogous to society' collective production of solved problems. Society faces, in a given period of time, some total number of problems, $|Tot|$, some portion of which---the set of solved problems in $\{PS\}$---are solved during that period of time. An individual's general problem solving capacity is therefore analogous to society's general collective problem solving capacity. If society's store of solved problems---its knowledge---increases, individual general problem solving capacity should increase. 

IQ researchers reason that it is not a change in the biology of human beings that leads to the increase in average IQs that has been measured by them in the past few decades. The human brain could not evolve that quickly.  Moreover, if intelligence increased that rapidly, then, extrapolating back in time, that would imply that the geniuses in the past, say 2000 years ago, achieved the results they did with meager intellectual resources, insufficient to accomplish what they did. It is therefore unlikely that the average innate, or basal, problem solving capacity of individuals has changed much over the past two or three thousand years. This has left researchers suggesting TV, diet, education, and so on, as possible factors that might explain why average IQs have been increasing.

Suppose that every person has an innate, or basal, general problem solving capacity. The innate, or basal, general problem solving capacity must be positive, otherwise it would not be possible for a person to figure out how to learn and remember the solution to a problem  solved by other people---to learn information. 

The few hundred problems that comprise an IQ test approximately measure the general problem solving skill of an individual. An individual's problem solving skill includes two components. One component depends on how well the person has learned the society's solved problems: how much knowledge the person has acquired from the social network and from society's store of solved problems. Another component depends on how much innate, or basal, problem solving capacity a person has---a capacity they were born with---to retrieve learned knowledge from that person's own memory and to process that learned knowledge---to vary, adapt and invent solutions to problems novel to the person using that knowledge. Comparing the average IQ for a group of people over a period of time measures a society's general collective problem solving rate by indirectly measuring the \textit{average} rate at which the \textit{effectiveness} of society's store of solved problems has increased, for all those individuals who have learned those solved problems. If the \textit{quality} of society's store of solved problems increases, the store of information of the socially networked individual should improve in proportion.  

\subsubsection{The average IQ and society's IQ}\label{Subsubsection deriving average IQs}

In this part we try to demonstrate that the rate of increase in the average person's general problem solving rate equals the rate of increase in society's average general collective problem solving rate. It would follow that the rate of increase of average IQs and the rate of increase of society's collective IQ should be equal. If that is so, we can validly use the rate of change in a society's average IQ as a means of estimating society's general, collective problem solving rate. 

First, we wish to demonstrate, with reference to IQs, that an average of a society's individual rates of problem solving is equivalent to the society's general collective problem solving rate, \textit{in principle}. 

We will use the following additional definitions, some of which are analogous to the problem solving definitions which begin on page \pageref{PSdefn}. 

\begin{IQdefinitions}\hfill
\begin{enumerate}
	\item 	$IQ_{Individual}(t) $: the IQ---intelligence---of an individual, as of time $t$.
	\item 	$IQ_{Av}(t) $: the average IQ---intelligence---of a set of individuals, as of time $t$. 
	\item 	$IQ_{Innate}(t)$: a theoretical construct, representing what an \textit{individual's} IQ would be without any social networking or language, as of time $t$. 
	\item 	$IQ_{Innate-Av}(t)$: a theoretical construct, representing what the \textit{average} individual's IQ would be without any social networking or language, as of time $t$. 
	\item $\{Networking_{Individual}(t) \}$: the set of solved problems of an individual that relate to social networking, as of time $t$.
	\item 	$f(Pop(t))$: a function of the population of a society, as of time $t$.
	\item 	$g (|\{PS_{Society}(t) \}|)$: a function of the number of a society's solved problems, as of time $t$.
	\item $[C(Pop)(t)]$ or $[C(Pop)]$: a proportion, between 0 and 1, of $f (Pop)$, as of time $t$. 
	\item $[C(PS_{Society} )(t)]$: a proportion, between 0 and 1, of $g (|PS_{Society}(t) |)$, as of time $t$. 
	\item For both $[C(Pop)(t)]$ and $[C(PS_{Society} )(t)]$, subscripts $Individual$ and, for the \textit{average} `$C$' value for a group of individuals, $Av$, are also used.
	\item $S-Pop(Ind)$: the base of the logarithmic function that measures part of the capacity of an individual to socially network.
	\item $S-PS(Ind)$: the base of the logarithmic function that measures part of the capacity of an individual to network with solved problems.
\end{enumerate}
\end{IQdefinitions}

\paragraph{The individual IQ.} In this first step, we relate \textit{an individual's} IQ to its component factors. 

Some proportion of $IQ_{Individual}(t)$ is based on the person's innate problem solving capacity---the problem solving capacity of a person determined by the person's genetically inherited, physiological capacity, $IQ_{Innate}(t)$. That is,

\begin{equation}\label{eq IQ-1.10.10}
\\IQ_{Individual}(t)\propto IQ_{Innate}(t).
\end{equation}

$IQ_{Individual}(t)$ is also proportional to a function of $| \{PS_{Individual}(t) \} |$, the amount of society's information---the number of solved problems that the individual has learned. The more an individual knows, the higher their working, individual IQ. The relationship between an individual's IQ and a function of the number of solved problems learned is linear. This must be the case for the following thermodynamic reasons. 

An individual, during their life, expends energy in an approximately linear way, proportional to time. 

\begin{equation}\label{eq IQ.energy 10.10}
( \Delta \ time)_{Individual} \propto (\Delta \ energy)_{Individual}.
\end{equation}

An individual, has the capacity to add to their existing personal store, or set, of solved problems, $\{PS_{Individual}(t) \}$, proportioned to the energy used by them to  acquire---learn---those solved problems, as follows: 

\begin{equation}\label{eq IQ.energy 10.20}
(\Delta \ energy)_{Individual} \propto \Delta g (|\{PS_{Individual}\}|).
\end{equation}

From (\ref{eq IQ.energy 10.20}), we infer

\begin{equation}\label{eq IQ.energy 10.24}
d \left[( energy)_{Individual}\right] \propto  d \left[g (|\{PS_{Individual}\}|)\right].
\end{equation}

The more that an individual knows---the more solved problems that the individual has learned---the more resources the individual has to use to solve newly encountered problems. So

\begin{equation}\label{eq IQ.energy 10.30}
 \Delta g(|\{PS_{Individual}\}|) \propto \Delta (IQ_{Individual}).
\end{equation}

Since an individual's use of energy is proportional to the advance of time, and the number of society's solved problems acquired by an individual is proportional to the energy used to learn them, it follows that the increase in an individual's store of solved problems is proportional to time. That implies that an individual's IQ increases linearly along with the linear increase in a function of the number of solved problems that the individual has learned. Hence, similarly to (\ref{eq IQ.energy 10.30}),

\begin{equation}\label{eq IQ-1.10.17}
\\ IQ_{Individual}(t)\propto g (|\{PS_{Society}(t)\} |).
\end{equation}

The portion of society's solved problems acquired---learned---by an individual is only a \textit{proportion} of a function of all of society's solved problems. That is, 

\begin{equation}\label{eq IQ-1.10.20}
	 g(|\{PS_{Individual}(t)\}|) \propto g (\{|PS_{Society}(t) \}|),
\end{equation}

\noindent or

\begin{equation}\label{eq IQ-1.10.25}
	 g(|\{PS_{Individual}(t)\}|) = [C(PS_{Society} )(t)]_{Individual} \times g (|\{PS_{Society}(t) \}|),
\end{equation}

\noindent where $[C(PS_{Society} )(t)]_{Individual}$---the `$C$' factor---represents the proportion of a function of all of society's solved problems. So, combining (\ref{eq IQ-1.10.17}) and (\ref{eq IQ-1.10.25}),

\begin{equation}\label{eq IQ-1.10.30}
	 IQ_{Individual}(t)\propto [C(PS_{Society} )(t)]_{Individual} \times g (|\{PS_{Society}(t) \}|).
\end{equation}

\noindent A solution to a problem that was solved by society can contain the solution to a problem that confronts an individual, or can help solve the individual's problem by analogy or by other methods. 

 From (\ref{eq IQ-1.10.10}) and (\ref{eq IQ-1.10.30}), it follows that, without regard to the effect of social networking, 

\begin{equation}\label{eq 22-1b}
\begin{split}
	 IQ_{Individual}(t) \propto  IQ_{Innate}(t) & \times [C(PS_{Society} )(t)]_{Individual} \\
& \times g (|\{PS_{Society}(t)\} |).
\end{split}
\end{equation}

\paragraph{Learned solved problems as a factor in IQ.} (\ref{eq 22-1b}) implies that an individual's capacity to solve problems, which is represented by $IQ_{Individual}(t)$ on the left side of (\ref{eq 22-1b}), arises from the individual applying their innate IQ to learn a portion---a subset---of the problems that society has solved. 

As to the innate IQ factor on the right side of (\ref{eq 22-1b}), the physiology and capacity of the average human brain has likely not changed much if at all over the past few thousand years. If a person's problem solving capacity is proportional to the physiological capacity of their brain, then it follows that the innate, or basal, problem solving capacity of the human brain---and the average individual innate IQ---has likewise remained  unchanged over the past few thousand years. Therefore, we assume that, at least in the recent past of human beings,

\begin{equation}\label{eq 22-3a}
\ \frac{d \left[IQ_{Innate}(t)\right]}{dt} = 0 .
\end{equation}

\noindent Based on all of these considerations, it must be that any rate of change in the number of solved problems that an individual learns is a result of applying $IQ_{Innate}$ to increase their store of solved problems. 

Since 

\begin{equation}
	IQ_{Individual}= \frac{d |\{PS_{Individual}\}|}{dt},
\end{equation}

\noindent we interpret (\ref{eq 22-1b}) as demonstrating that an individual IQ test score is one possible metric for measuring the problem solving capacity, or IQ, of an individual that arises when an individual's innate, or basal, problem solving capacity, or innate IQ, is applied to a subset of society's solved problems. An individual's innate, or basal, problem solving capacity, or innate IQ, is also involved in solving the problem of \textit{how to learn} that subset of society's solved problems. An individual's store of solved problems cannot increase without the individual applying their innate, or basal, problem solving capacity; without an individual invoking their innate, or basal, problem solving capacity, the individual's store of solved problems remains static. (\ref{eq 22-1b}) suggests that the innate IQ on the right side of (\ref{eq 22-1b}), as an \textit{invented, notional, metric}, measures, at least partly,  an individual's problem solving capacity, which is equivalent to the processing part of the left side of (\ref{eq 22-1b}). 

This relationship is analogous to a relationship in thermodynamics, between energy and absolute temperature. \textit{Absolute temperature} is an \textit{invented metric}---and is \textit{not} a function of time---such that the absolute temperature is, for a given closed system, proportional to the system's heat energy. If the heat energy in the system doubles, the absolute temperature doubles. By measuring the absolute temperature, one can indirectly measure any increase in the given system's heat energy content. The ratio of heat energy to absolute temperature for a given closed system is its entropy, which can be described as the ratio

\begin{equation}\label{eq eta-10.20}
\begin{split}
	 Q / T &= \eta\\
&= log_{S} (W),
\end{split}
\end{equation}

\noindent where $Q$ is the heat energy of the system, $ \eta $, the small Greek letter eta, represents the entropy of the system, $S$ is the base of the logarithmic function, $T$ is the absolute temperature of the system, and $W$ can be considered to be the state of the system, or the number of individual energy components in the system \cite[p. 118, for example]{Planck1914}. (There is a constant $k$ that multiplies $\log_{S} (W)$, but we will suppress that for now, as if $k=1$.)

Absolute temperature, as a source for our analogy, and an individual IQ test score, as a target of an analogy, does not work exactly. As was mentioned above, the standard IQ test score, based on a metric of 100 as the average IQ, is adjusted over time; the standard average 100 IQ is \textit{not} absolute. In the target of our analogy, for a given individual and also for the average individual, the number of solved problems per IQ point \textit{increases} over time, while for a \textit{given thermodynamic system}, the amount of energy per degree of temperature is unchanging. 

The problem solving capacity of an individual as it is measured by an IQ test score is affected by education---how much the individual has learned, as implied by (\ref{eq 22-1b}). If, having regard for (\ref{eq 22-1b}), we treat problem solving \textit{work} as being the result of an individual applying $IQ_{Innate}$ using what the individual has learned, then if we want to focus on a problem solving capacity of an individual without the enhancement of education, we should use the theoretical, or notional, unchanging $IQ_{Innate}$ of an individual as a metric of that individual's problem solving capacity, analogous to absolute temperature as a metric of heat content for a given system. 

For an individual, $IQ_{Innate}$ is unchanging during their life, and so, for that individual, is analogous to absolute temperature, which is an unchanging metric for a thermodynamic system. Similarly, $IQ_{Innate-Av}$ for a set of people---the average innate IQ for that set of people---is similarly an unchanging metric for their average problem solving capacity, or rate. If the average innate IQ has remained relatively constant over the past few thousand years, then it may be considered to be more analogous to the idea of absolute temperature than is individual (composite) IQ which changes, and as well, is partly a function of what a person learns. The idea of an innate IQ is entirely theoretical, because it would be difficult in practice to isolate a person's innate problem solving capacity from what the person has learned.

Let's consider how an innate problem solving capacity of an individual interacts with \textit{a network of solved problems}. To simplify our analysis, and consistent with the foregoing, we first assume that this occurs in the absence of a social network. By way of an \textit{approximate} analogy to (\ref{eq eta-10.20}),

\begin{equation}\label{eq eta-10.22}
	\frac{ d \left|\left\{PS_{Individual}(t)\right\}\right|}{dt} \equiv Q
\end{equation}

\noindent (the change in information is analogous to energy, because new information is created by the input of energy, and because as the number of solved problems increases, the problem solving rate increases),

\begin{equation}\label{eq eta-10.24}
	IQ_{Innate}(t) \equiv T
\end{equation}

\noindent (the invented IQ metric, but at the level of the notional $IQ_{Innate}$, is analogous to absolute temperature), and

\begin{equation}\label{eq eta-10.30}
\begin{split}
	 & \frac{ d \left|\left\{PS_{Individual}(t)\right\}\right|}{dt} / IQ_{Innate}(t) \\
&= \eta\\
&= \log_{S-PS(Ind)} (\left|\left\{PS_{Individual}(t)\right\}\right|), 
\end{split}
\end{equation}

\noindent (as in (\ref{eq eta-10.20}), forming an analog to a definition of entropy used in thermodynamics), where the base of the logarithm, $S-PS(Ind)$, is applicable, for a particular individual, to a network of solved problems---what might called a network of abstractions or ideas.

From (\ref{eq eta-10.30}), and multiplying the left and right sides by $IQ_{Innate}(t)$, we have

\begin{equation}\label{eq eta-10.25}
\begin{split}
	 \frac{ d \left|\left\{PS_{Individual}(t)\right\}\right|}{dt} &= IQ_{Innate}(t) \\
& \times \log_{S-PS(Ind)} \times (\left|\left\{PS_{Individual}(t)\right\}\right|) .
\end{split}
\end{equation}

The rate of change, on the right side of (\ref{eq eta-10.25}), in the number of solved problems learned by a problem solving individual, arising from the application of an individual's innate IQ, is equal to the individual's composite intelligence, $IQ_{Individual}(t)$, since, for the left side of (\ref{eq eta-10.25}),

\begin{equation}\label{eq eta-10.40} 
	 \frac{ d \left|\left\{PS_{Individual}(t)\right\}\right|}{dt} = IQ_{Individual}(t).
\end{equation}

It follows from (\ref{eq 22-1b}) and (\ref{eq eta-10.40}), now having regard for the \textit{proportion} $C(PS)$ of $\log_{S-PS}(|\{PS_{Society}(t)\}|)$ that an individual knows, that

\begin{equation}\label{eq eta-10.50}
\begin{split}
\  \frac{ d \left|\left\{PS_{Individual}(t)\right\}\right|}{dt} &=   IQ_{Innate}(t) \\
&  \times [C(PS_{Society} )(t)]_{Individual} \\
& \times \log_{S-PS}(|\{PS_{Society}(t)\}|) .
\end{split}
\end{equation}

In the second and third lines in (\ref{eq eta-10.50}), $[C(PS_{Society} )(t)]$ and $|\{PS_{Society}(t)\}|$ likely change over the course of a person's lifetime. $|\{PS_{Society}(t)\}|$ changes as the result of collective problem solving; $[C(PS_{Society}(Ind) )(t)]$ can change as a result of additional energy spent by an individual in learning solved problems---increasing the proportion of society's solved problems that the individual has learned. 

The factor 

\begin{equation}\label{eq eta-10.60}
	 \left[C(PS_{Society}(t))\right]_{Individual} \times \log_{S-PS}(|\{PS_{Society}(t)\}|)
\end{equation}

\noindent in (\ref{eq eta-10.50}) exactly corresponds to the general form that entropy takes in thermodynamics, $C \times log_{S} (n)$. 

The situation represented by (\ref{eq eta-10.50}) is a theoretical, or notional, one in which individuals have direct access to a store of society's solved problems, but \textit{without any social networking}. 

The implication of (\ref{eq eta-10.50}) is that the factor set out in (\ref{eq eta-10.60}) multiplies innate IQ, and as well, that the more `solved problem' targets an individual's innate IQ can network with, the more `intelligent' that individual will be.

\paragraph{Social networking as a factor in IQ.} We now consider the effect of social networking on $IQ_{Individual}(t)$, an individual's IQ. For the purposes of this article, I accept as true the following observations:

\begin{quotation}
	``As the child negotiates the early developmental levels through countless emotional exchanges with her caregiver, she develops an implicit understanding of her society's attitudes towards beliefs and social practices, norms and values, power hierarchies and the kinship system, and so on \cite[p. 325]{Greenspan2004}. 

\ldots it is the pattern of reciprocal, co-regulated affective interactions in the early stages of development that helps the child differentiate her own individual personality and helps the group determine its collective personality'' \cite[p. 332]{Greenspan2004}.
\end{quotation}

% THE ENTROPY OF THE SOCIAL NETWORK AS A FACTOR

 How to socially network is a problem solving exercise, beginning with an individual's earliest childhood moments. Each `emotional exchange' requires each party to the exchange to solve the \textit{problem} of how to participate in that exchange. A child has to decipher her caregiver's signals, and has to solve the problem of how to signal back to the caregiver. `All life is problem solving.'
 
 Analogous to the way that we considered the thermodynamics of how an individual \textit{networks with a society's solved problems and abstractions}, let's consider \textit{how an individual networks with other individuals in their society}---how the individual solves the problem of how to socially network---but ignoring, for now, society's network of solved problems. We can infer that the capacity of an individual to socially network is analogous to their capacity to solve \textit{social networking problems}.  We can consider the \textit{set} of individuals with whom a given individual networks, or equivalently, the \textit{set} of solved problems,  corresponding to `solved' problems that establish social relationships, $\left\{Networking_{Individual}(t)\right\}$, to consist of a population connected, directly and indirectly, to an individual. That is 

\begin{equation}\label{eq Networking-10.10}
	 \left\{Networking_{Individual}(t)\right\} =\left\{Pop_{Individual}(t)\right\}.
\end{equation}

$\left|\left\{Pop_{Individual}(t)\right\}\right|$ represents the person's social network. Since an individual must solve problems of how to network with other individuals, by analogy to (\ref{eq eta-10.30}),
\begin{equation}\label{eq Networking-10.30}
\begin{split}
	 & \frac{d\left|\left\{PS_{Networking}(t)\right\}\right|}{dt} / IQ_{Innate}(t) \\
&= \eta\\
&= \log_{S-Pop(ind)} (\left|\left\{Pop_{Individual}(t)\right\}\right|). 
\end{split}
\end{equation}

\noindent By analogy to (\ref{eq eta-10.50})
\begin{equation}\label{eq Networking-10.60}
\begin{split}
  & \frac{ d \left|\left\{Networking_{Individual}(t)\right\}\right|}{dt}\\
&=   IQ_{Innate}(t)  \times [C(Pop)(t)]_{Individual} \times \log_{S-Pop}(Pop(t)) ,
\end{split}
\end{equation}

\noindent where ${S-Pop}$ is the base of the logarithmic function applicable for the whole society.

In light of (\ref{eq 22-1b}), and since an individual's social network arises from the application of $IQ_{Innate}$, any rate of change that a person experiences in their social network ultimately is a result of applying $IQ_{Innate}$ to the problems of how to socially network with other individuals. It must be therefore, that, similarly to (\ref{eq eta-10.40}),

\begin{equation}\label{Networking-10.50}
	 \frac{ d \left|\left\{Networking_{Individual}(t)\right\}\right|}{dt} \propto IQ_{Individual}(t).
\end{equation}

The situation described by (\ref{eq Networking-10.60}) is a theoretical, or notional, one in which individuals socially network, but \textit{do not} network with solved problems. 

The factor 

\begin{equation}\label{eq Networking-10.70}
\\ \left[C(Pop)(t)\right]_{Individual} \times \log_{S-Pop}(Pop(t))
\end{equation}

\noindent in (\ref{eq Networking-10.60}) also corresponds to $C \times log_{S} (n)$, the general form that entropy takes in thermodynamics.

Now we suppose that an individual can contemporaneously network with other individuals, a situation described by (\ref{eq Networking-10.60}), and with a set of solved problems, a situation described by (\ref{eq eta-10.50}). If we blend together the two situations described by (\ref{eq eta-10.50}) and (\ref{eq Networking-10.60}), we obtain a combined formula that describes individual intelligence, as follows: 

\begin{equation}\label{eq Networking-10.1000}
\begin{split}
 & \frac{ d \left|\left\{PS_{Individual}(t)\right\}\right|}{dt}\\
 &=  IQ_{Individual}(t) \\
&=   IQ_{Innate}(t) \\
& \times [C(PS_{Society} )(t)]_{Individual} \times \log_{S-PS}(|\{PS_{Society}(t)\}|) \\
& \times \left[C(Pop)(t)\right]_{Individual} \times \log_{S-Pop}(Pop(t)) .
\end{split}
\end{equation}

The three components, or factors, of an individual's IQ in (\ref{eq Networking-10.1000}), appear to be:

\begin{itemize}
	\item the individual's innate IQ;
	\item a function of the individual's knowledge;
	\item a function of the individual's social network.
\end{itemize}

\noindent There is likely a fourth factor that affects individual intelligence, represented by the combined effect of the environment, culture and infrastructure, to the extent not already subsumed within the three components just mentioned.  

With respect to (\ref{eq Networking-10.1000}) the population $(Pop(t))$ of a society can be estimated. It is possible, if \textit{The General Collective Problem Solving Capacity Hypothesis} is valid, to estimate the number of solved problems $|\{PS_{Society}(t)\}|$ by finding an enumerated set of collectively solved problems proportional to that number. That leaves us with the problems of measuring, for an individual, the other four parameters in (\ref{eq Networking-10.1000}), namely $C$ and $S$ for each of society's population and its solved problems. $[C(PS_{Society} )(t)]_{Individual}$ represents that proportion of society's solved problems that an individual has `networked' with. $S-PS$, the base of the logarithm of the number of society's solved problems, must have some thermodynamic relationship to society's store of solved problems due to its relationship to entropy. $[C(Pop )(t)]_{Individual}$ may represent how well an individual is socially networked, or how adept at social networking an individual is. $S-Pop$, the base of the logarithm of society's population, must also have some thermodynamic relationship to society's population. These four parameters, the two $C$'s and the two $S$'s in (\ref{eq Networking-10.1000}), may be difficult to measure for an individual, but they have been measured as averages for nodes in some networks.

Later in this article we discuss the problem of how to characterize and measure the following four parameters in (\ref{eq Networking-10.1000}):

\begin{itemize} \label{items 4qs of C and S}
	\item $[C(PS_{Society} )(t)]_{Individual}$
	\item $S-PS$
	\item $[C(Pop )(t)]_{Individual}$
	\item $S-Pop$.
\end{itemize}

\paragraph{Average IQs.} We can evaluate the rate of change in the \textit{average} individual IQ in terms of its component factors, by using (\ref{eq Networking-10.1000}). 

Let the subscript $i$ in the following be used to identify the different individuals comprising the society under consideration. For a society consisting of $n$ individuals, the average of the society's individual IQs is: 

\begin{equation}\label{eq 23}
\\ IQ_{Av} = \frac{\sum^{n}_{i=1}\left[IQ_{individual}\right]_{i} } {n}.
\end{equation}

To simplify notation for the next equation, let the following represent the indicated values, all at a common time $t$.
\begin{itemize}
	\item $[C(PS_{Soc} )]_{i}$: the proportion of the number society's store of solved problems learned by individual $i$;
	\item  $|\left\{PS_{Soc}\right\}|$: the number of society's store of solved problems;
	\item $[S-PS](i)$: the base of the logarithm for the logarithmic function of the number of society's solved problems, for individual $i$; 
	\item $\left[C(Pop)\right]_{i}$: the proportion of the society's network connected to individual $i$;
	\item Pop: the population of the society;
	\item $[S-Pop](i)$:  the base of the logarithm for the logarithmic function of the society's population, for individual $i$.
\end{itemize}

Using this notation, and combining the results of  (\ref{eq Networking-10.1000}) and (\ref{eq 23}), we obtain, letting the subscript $Av$ represent the average for the society's individuals, at a time $t$, 

\begin{equation}\label{eq 24}
\begin{split}
 IQ_{Av}(t)
&= \frac{\sum^{n}_{i=1}\left[IQ_{Individual} \right]_{i} } {n}\\
 & = [ \ \sum^{n}_{i=1} \left[IQ_{Innate}\right]_{i} \\
 & \times [C(PS_{Soc} )]_{i} \times \log_{[S-PS](i)}(|\{PS_{Society}\}|) \\
 & \times \left[C(Pop)\right]_{i} \times \log_{[S-Pop](i)}(Pop)  ] \\
 & \times \frac{1}{n} .
\end{split}
\end{equation}

We infer that the averaged values of the two $C$s and $S$s that help determine the IQ of the average individual in a society are related to the $C$s and $S$s with the subscript $i$ that apply to individuals in (\ref{eq 24}). Consistent with our hypothesis about the way a general problem solving capacity works, the same \textit{general} problem solving capacity applied to learning \textit{society's problems} should apply to solving problems of how to socially network. Consistent with \textit{The General Collective Problem Solving Capacity Hypothesis}, each component factor in (\ref{eq 24}) achieves its average value in a similar way. In other words, the average of the product on the right side of (\ref{eq 24}) should equal the product of the \textit{averages}.  So we infer, based on (\ref{eq 24}), that, at a time $t$

\begin{equation}\label{eq 24-Av}
\begin{split}
 IQ_{Av}
 &= \frac{\sum^{n}_{i=1}\left[IQ_{Individual} \right]_{i} } {n}\\
&= \frac{\sum^{n}_{i=1}\left[IQ_{Innate} \right]_{i} } {n}\\ 
 & \times [C(PS_{Soc} )]_{Av} \times \log_{[S-PS](Av)}(|\{PS_{Society}\}|) \\
 & \times \left[C(Pop)\right]_{Av} \times \log_{[S-Pop](Av)}(Pop) \\
 & =  \left[IQ_{Innate}\right]_{Av} \\
 & \times [C(PS_{Soc} )]_{Av} \times \log_{[S-PS](Av)}(|\{PS_{Society}\}|) \\
 & \times \left[C(Pop)\right]_{Av} \times \log_{[S-Pop](Av)}(Pop) .
\end{split}
\end{equation}

\noindent (\ref{eq 24-Av}) should apply for any rate of population growth, including a rate of zero population growth. For the past few thousand years, a human being's genetically endowed problem solving capacities have been unchanged, which implies that the average innate, or basal, IQ has been unchanged during that period of time, consistent with (\ref{eq 22-3a}). If we take the derivative with respect to time of the left side of (\ref{eq 24-Av})---average IQ---for a period in which there is no population growth, and of the product on the right side of (\ref{eq 24-Av}), we have 

\begin{equation}\label{eq 24-2b}
\begin{split}
	 \frac{d \left[IQ_{Av}(t)\right]}{dt}& = \left[IQ_{Innate}\right]_{Av} \\
& \times \frac{d \left([C(PS_{Soc} )]_{Av} \times \log_{S-PS(Av)}(|\{PS_{Society}\}|) \right) }{dt},
\end{split}
\end{equation}

\noindent because $IQ_{Innate}$, as in (\ref{eq 22-3a}), does not change.

In other words, (\ref{eq 24-2b}) predicts that the average rate at which average IQs increase, for a period during which the population size is relatively unchanging, should equal  the average \textit{innate} IQ times the average rate at which the logarithmic function---the entropy---of an enumerated set of collectively solved problems increases. 

\paragraph{Society's IQ} Analogously to the situation of an individual, a society's IQ is proportional to its store of solved problems, considering society as if it were one (networked) individual having a `single mind' in itself: 

\begin{equation}\label{eq Soc-IQ.10}
\\ IQ_{Society} \propto g(|\{PS_{Society}\}|).
\end{equation}

Not all solved problems in a society are perfectly networked, but only a proportion, $C(PS_{Soc} )$. That is, 

\begin{equation}\label{eq Soc-IQ.20}
	 IQ_{Society} \propto C(PS_{Soc} ) \times g(|\{PS_{Society}\}|).
\end{equation}

If we take the derivative with respect to time of the left and right sides of (\ref{eq Soc-IQ.20}), then we have

\begin{equation}\label{eq Soc-IQ.30}
	 \frac{d \left[IQ_{Society}\right]}{dt} \propto \frac{d \left([C(PS_{Soc} )] \times \log_{S-PS}(|\{PS_{Society}\}|)\right) }{dt}.
\end{equation}

The proportion $C(PS_{Soc} )$ for the whole society equals $C(PS_{Soc} )_{Av}$. Similarly, for the whole society, $[S-PS] = [S-PS]_{Av}$. Therefore,

\begin{equation}\label{eq Soc-IQ.30Av}
\begin{split}
	 \frac{d \left[IQ_{Society}\right]}{dt} &= \left[IQ_{Innate}\right]_{Av} \\
& \times \frac{d \left([C(PS_{Soc} )]_{Av} \times \log_{S-PS(Av)}(|\{PS_{Society}\}|) \right) }{dt}.
\end{split}
\end{equation}

\noindent The right side in (\ref{eq Soc-IQ.30Av}) equals the right side in (\ref{eq 24-2b}), and so,

\begin{equation}\label{eq Soc-IQ.40}
	 \frac{d \left[IQ_{Society}\right]}{dt}= \frac{d \left[IQ_{Av}(t)\right]}{dt}.
\end{equation}

Another way of obtaining the result in (\ref{eq Soc-IQ.40}) is to observe that society's IQ---its problem solving capacity---consists of the \textit{unnetworked} average problem solving capacities of its individual members, times $C(Pop) \times \log_{S-Pop}(Pop)$. So

\begin{equation}\label{eq 22-8}
	IQ_{Society}(t) = IQ_{Av}(t) \times C(Pop) \times \log_{S-Pop}(Pop). 
\end{equation}

\noindent If we substitute into (\ref{eq 22-8}) the value of $IQ_{Av}$ from (\ref{eq 24-Av}), and take the derivative of the left and right sides, for a society in which there is no population growth, we obtain the result in (\ref{eq Soc-IQ.40}).

One implication of (\ref{eq 22-8}) is that if society's intelligence increases, then  average  IQs will increase in proportion, or, 

\begin{equation}\label{eq 22c}
\\IQ_{Society} \nearrow \ \Longrightarrow IQ_{Average}\nearrow.
\end{equation}

\noindent If average IQs increase, that necessarily implies that for most everyone in society,

\begin{equation}\label{eq 22c-2}
\\IQ_{Society} \nearrow \ \Longrightarrow IQ_{Individual}\nearrow.
\end{equation}

The problem solving capacity of an individual in a society is greater than the problem solving capacity of the individual without that society, an observation implicit in (\ref{eq 24-Av}). The more people there are, the more problem solving resources there are: more memory and more processing power. Society has more networked neurons than an individual does. \textit{How much greater are the problem solving resources of society than those of the average individual?} 

We have demonstrated in (\ref{eq Soc-IQ.40})

\begin{Society's IQ Theorem} The rate of growth in society's collective IQ equals the rate of growth in the average individual IQ.
\end{Society's IQ Theorem}

\textit{Society's IQ Theorem} allows us to obtain a rate for society's general collective problem solving rate by using statistics about the rate of change in \textit{average} individual IQs. 

The \textit{general} problem solving capacity of a person is analogous to the \textit{general}, collective problem solving capacity of society.

\section{Data, calculations and measurements for the general problem solving capacity hypothesis}\label{Section data calcns}

I propose to use three sources of data to estimate society's average general collective problem solving rate:  the rate of increase in average IQs,  the size of the English lexicon, and the improvement in lighting efficiency. In these data, \textit{the proportion of the total population} involved in, or measured for, their, evaluative problem solving, and \textit{the number} of problems solved, should both be large enough to permit us to reasonably estimate society's average general collective problem solving rate. Because of the range and generality of problems posed respectively in IQ tests, in evaluating words in the lexicon, and in evaluating lighting improvements, I infer that each of the three kinds of data enable measurements of the rate of solving problems that are representative of society's general collective problem solving rate. 

\subsection{Increasing average IQs}\label{Subsection IQ data}

The first set of data we consider concerns the rate at which average IQs increase.\begin{quote}
	``Reed Tuddenham was the first to present convincing evidence using a nation-wide sample''\cite[p. 2]{Flynn2007}
\end{quote}

\noindent that average IQs are increasing. In this part, we use the average rate of increase in average IQs as determined by IQ researchers. The  average rate of increase in average IQs is equal to the average rate of increase in society's IQ, by reason of \textit{Society's IQ Theorem}, giving us a statistic that measures the average general collective problem solving rate. 

The rate of a society's increase in average IQs is estimated by comparing average IQ test scores at different times. The average increase in IQ test scores based on compared average IQ test scores, on a standardized basis, for the United States from 1947 to 2002, is $0.300$ to $0.363$ IQ points per year \cite[p. 113 and Table 1 at p. 180]{Flynn2007}. The research results describe a \textit{linear} rate of increase of 3.00 to 3.63 IQ points \textit{per decade}. Since the cumulative number of society's solved problems should in principle grow exponentially over time, and since it is  difficult to distinguish the difference between a \textit{linear} rate of growth and an \textit{exponential} rate of growth when the growth rate is slow over a short period of time, which is the case for this average IQs data, I infer that the rate of increase in average IQs must be \textit{exponential}.

Since a 100 IQ is the mean, we can infer that the \textit{exponential} rate of increase in average IQ test scores is about 3.00\% to 3.63\% per decade; the midpoint of those rates is about 3.315\% per decade. The 0.63 difference between the high and low measurements of the rate of increase in average IQs is about 17.35\% of the higher measurement of 3.63\% per decade. This spread in measurement is less than, but consistent with, the descriptions of the percentage errors, summarized in Tables \ref{Table 1-1} and \ref{Table 1-1Lower}, that can arise due to a 10\% error in estimating $N(t_2)$ over a short period of time, such as a period of several decades. Still, we can consider this estimate, 3.315\% per decade, as set out in Table \ref{Table 1.1.2}, to be an approximation of the actual rate at which average IQs increase. 

\renewcommand{\arraystretch}{1.25}
\begin{table}[ht]
\begin{center}
	\begin{tabular}{|c|c|}\hline
	Years&The average rate of increase in average IQs\\ \hline
	1945--2002&3.315\% per decade\\\hline
	\end{tabular}
	\caption{The estimated average rate of increase in average IQs.}\label{Table 1.1.2}
	\end{center}
\end{table}

\subsection{Language, and its lexical growth rates}\label{Subsection lexical data}

If \textit{The General Collective Problem Solving Capacity Hypothesis}---the hypothesis that society possesses a \textit{general} problem solving capacity---is valid, then the rate at which a society solves language problems \textit{in particular}, should be the same as the society's \textit{general} collective problem solving rate. Which is to say that it would be the case that

\begin{equation}\label{eq 2b}
\\ \frac{d |\{PS_{Lang} (t)\}|  }{dt} = \frac{d |\{PS_{Society} (t)\}| }{dt}.
\end{equation}

Language is also a medium that \textit{helps} solve problems that can be expressed conceptually---by using abstractions. Since language helps solve problems, language must improve some aspect of the process of problem solving that leads to an increase in the number of solved problems; otherwise language would not play such a prominent role in human society. \textit{How quantitatively important is language's role in human problem solving}? \textit{How much more intelligent is a human being with language than without it?} are questions that follow from considering these aspects of language.

To use language in problem solving requires identifying, manipulating and combining appropriate abstractions. An abstraction encoded by sound, that quickly communicates danger or opportunity over distances, around corners and in the dark, is useful.  Combining different sounds increases society's capacity to encode different abstractions. If there are too many sounds, though, it would impede learning them all. 

Words, as reified concepts, can be \textit{metaphorically} manipulated: we `make' and `build' arguments, and `construct' theories. Manipulating worded abstractions to solve a problem conceptually is analogical to manipulating physical objects---solving the problem physically---if the abstractions accurately encode and model the corresponding real world phenomena. Since manipulation of physical objects is familiar to people, manipulation of words---things represented by encoded abstractions---becomes familiar by a kind of analogy: the relationship between things is modeled by the relationship between the corresponding abstractions.

Since language increases a person's problem solving capacity, which is beneficial for society, society continuously seeks ways to make language more efficient. Language itself encompasses different kinds of problems that require solving. If \textit{The General Collective Problem Solving Capacity Hypothesis} applies to language problems, the different kinds of language problems should themselves be solved at the same collective rate. 

Members of society continuously, collectively, solve language problems that include:  

\begin{enumerate}
	\item How should society invent, choose and improve a set of sounds for encoding abstractions?
	\item Which phenomena can be abstracted? 
	\item Which phenomena, for which abstraction is possible, are useful to abstract?
	\item How, using the chosen set of sounds, should the chosen phenomena be encoded? 
	\item How efficient is an abstraction?
	\item How efficient are procedures for organizing abstractions, such as rules of grammar?
	\item Which words and language procedures, such as rules of grammar, should be improved, preserved or discarded?
\end{enumerate}

\noindent The current state of a language stores society's solutions to these language problems. 

Solved language problems are an information resource for the society speaking the language. Solved language problems are an emergent infrastructure of encoded abstractions that assists in solving new language problems, by analogy, variation, combination, and (adopting the term used by Fauconnier and Turner in \cite{Fauconnier2002}) blending, all created by networked individuals over many generations.

The lexicon of a language improves by increasing the \textit{depth} of words---increasing the amount of information a word connotes---and by increasing the \textit{range} of words---adding new words to the lexicon. 

Increasing the \textit{depth} of words and grouped words is analogous to increasing the amount of information in a transmitted signal by compressing the encoded data.  In language, information compression is achieved in various ways, by naming, by using abbreviations, initials,  shortening the length of words---for example, from `internetwork' to `internet'---and by using allusive words and phrases to which the society's culture and general knowledge supply additional information. Examples of allusive compression are `WWII' and `personal computer.' A similar kind of allusive compression of abstraction is exhibited by mathematical notation, such as the symbol for integration, `$\int$,' and by mathematical conventions, such as the order  in which binary operations are performed.

If society can increase the level of compression of information in words---increase the \textit{depth} of words---the effect on the capacity of individuals and their societies to solve problems should be, over time, dramatic. If ten compressed words contain twice as much information as their ten predecessor words, then the problem solver can now manipulate twice as much information as before, in the same amount of time. The problem solver's problem solving capacity with respect to the manipulation of those encoded abstractions has increased (logarithmically with respect to those ten words).\footnote{That is, by a factor of $\log_S(2)$, when the 10 predecessor words are compared to the 10, twice as compressed, successor words.} Since the lifetime of a person is finite, society's capacity to increase the \textit{depth} of  words \textit{increases} the amount of information that can be manipulated within the same amount of time and with the same amount of effort, and can, and I believe does, increase the problem solving capacity of an individual. 

It is difficult to devise a way to measure the effect of increasing \textit{compression} on the efficiency of language. \textit{The General Collective Problem Solving Capacity Hypothesis} implies, however, that if we find some \textit{other feature} of language that is enumerable, then the rate of increase in that \textit{enumerable} language feature should correspond to the rate at which \textit{compression} improves the \textit{depth} of abstractions in a lexicon. 

A second category of language problems can be considered to result in the increase of the \textit{range} of a lexicon, or the number of words in a language.  I am not sure how the rate of change in the \textit{depth} of words could be directly measured but as to the \textit{range} of words, we can count words in a dictionary. I propose to consider, as a proxy for the measurement of the increasing compression---the increasing \textit{depth}---of language, the increasing \textit{range} of words in a lexicon. The range of words in a language is the number of words, and the number of words in a lexicon \textit{is} a countable feature of a language. 

If \textit{The General Collective Problem Solving Capacity Hypothesis} is valid, a society's average \textit{lexical growth rate} should, in principle, equal the rate of increase in the \textit{depth of compression} of words. The average rate of increase in the \textit{depth} and \textit{range} of words should equal the society's \textit{average} \textit{language problem} solving rate; a society's average \textit{language} problem solving rate should, in principle, equal the society's average general collective problem solving rate---society's problem solving capacity---because \textit{The General Collective Problem Solving Capacity Hypothesis} implies that all these equalities hold.

In effect,

\begin{equation} \label{eq 3-2.20}
	 \left\{PS_{Depth}\right\} \subset \left\{PS_{Lex}\right\},
\end{equation}

\begin{equation} \label{eq 3-2.40}
	 \left\{PS_{Range }\right\} \subset \left\{PS_{Lex}\right\} ,
\end{equation}

\noindent and

\begin{equation} \label{eq 3-2.50}
	 \left\{PS_{Lex}\right\} \subset \left\{PS_{Lang}\right\} \subset \left\{PS_{Society}\right\} .
\end{equation}

\noindent It would follow that, for language, 

\begin{equation} \label{eq 3-2.60}
\begin{split}
\ \frac{d \left|\left\{PS_{Depth}\right\}\right|}{dt}&=\frac{d\left|\left\{PS_{Range }\right\}\right|}{dt}\\
&= \frac{d\left|\left\{PS_{Lex }\right\}\right|}{dt}
\end{split}
\end{equation}

and

\begin{equation}\label{eq 3-2.70}
\begin{split}
\frac{d\left|Lex(t)\right|}{dt} 
& = \frac{d|\{PS_{Lang}(t)\}|}{dt} \\
& =\frac{d|\{PS_{Society}(t)\}|}{dt} .
\end{split}
\end{equation}

We have converted the vague problem of finding the average general collective problem solving \textit{capacity} of society to one of finding a representative problem solving \textit{rate}---an average lexical growth \textit{rate}.  If average problem solving rates for society's other general collective problems coincide with the average lexical growth rate, then we will have evidence that society has an average general collective  problem solving \textit{rate}---and hence, an average, general collective problem solving \textit{capacity}.

The second set of data we consider in this section concerns average English lexical growth.  For the English lexicon, I use the number of words in the 1989 Oxford English dictionary, and data from two University of Toronto (U of T) historical dictionary projects, to estimate the size of the English lexicon for different years. To find society's average general collective problem solving rate requires the use of  (\ref{eq 10}). We wish to show that the English lexicon grows at the same rate as English society's average general collective problem solving rate, because the number of problems solved in the lexicon is proportional to the number of general collective problems solved by society, that is, 

\begin{equation}\label{eq 21}
\\ Lex(t)=K(t) \times \left[PS_{English \ Society}(t) \right],
\end{equation}

\noindent where $K(t)$ represents the proportion. It is possible that $K(t)$ could vary over time.

A lexicon results from thousands, perhaps millions of people,  over many generations, solving evaluative problems about the convenience of the \textit{components of language}, and about the efficiency of the networking of such components. Thus we have a large group of people confronted daily with lexical problems, and a large problem set, including the problems of devising, refining, evaluating and improving  the set of sounds to be used for encoding, how to encode abstractions into words, how to group words into more complex abstractions, and how to store the accumulation of encoding solutions. The components of language, including the lexicon, are a much larger sampling of a society's average, general, collectively solved problems than are the questions on an IQ test. Almost everyone in society, a larger group of people than those few who take IQ tests and those fewer who design them, is involved in teaching, learning, and providing feedback about the efficiency---convenience---of words and language. With respect to the English lexicon, the daily evaluation by millions of people of thousands of words is more frequent then the occasional, if ever, evaluation of the problem solving capacity of those people by IQ testing. The English lexicon provides a larger  set of solved problems, that have been evaluated by more people, more frequently, over a historically longer period of time, than IQ tests. 

What counts as a word requires a professional judgment. Likely, the OED and U of T's two historical English dictionary projects, relied on in this part, use relatively consistent criteria for what counts as a word. We make the necessary assumption that the criteria for what counts as a word are similar for U of T's Old English project, U of T's Early Modern English Dictionaries Database (EMEDD), and the OED. Historical dictionaries may miss words that were used in speech but not recorded; that may be partly counter-balanced by inclusion in a dictionary of words rarely used. For data, I use three sources for English lexical data. 

The first source of English lexical data is the University of Toronto's partly completed Old English Dictionary \cite{DOE}. It covers the period of Old English, from the year 600 to the year 1150. Eight of  22 Old English letters, up to the letter \textit{g}, have been completed.\footnote{at December 2008} Based on, and extrapolating from the total of 12,271 words for the 8 completed letters---the dictionary counts \ae\ as a separate letter---and assuming the same average number of words per letter, I estimate  34,020 words in Old English for the whole Old English alphabet of 22 letters. If the number of words per letter for the next, uncompleted, 14 letters of the Old English alphabet is different than the average number of words for the dictionary's first eight letters, then the estimate may be slightly in error. If the estimate of the average number of words per letter, based on the first 8 letters,  is valid, it is likelier that the estimate of the total number of words is low rather than high; a dictionary that ends in the year 1150 omits Old English words that were used but never written down. 

The  most complete previous compilation of Old English words is \textit{An Anglo-Saxon and English Dictionary} \cite{Bosworth1898}, which has a  1921 supplement \cite{Toller1921}, revised in 1972 \cite{Campbell1972}, and covers the period 450 to 1100. Likely, H. Gneuss relies on \textit{An Anglo-Saxon and English Dictionary}  for his estimate of about 30,000 recorded Old English words,``although there must be gaps \ldots'' \cite{Gneuss1991}.\footnote{Dr. Oliver M. Traxel of the University of Muenster referred me to Gneuss's article in September 2005.}

The second source of English lexical data is another University of Toronto project, the Early Modern English Dictionaries Database \cite{EMEDD}. Information about it can be found on the internet. As of 1999, when the EMEDD project concluded, it had about 200,000 word-entries to the year 1657. It is not unlikely that the EMEDD omits some words never recorded. On the other hand, some words in the EMEDD had ``foreign language headwords, and some headwords were repeated.''\footnote{November 26, 2008 personal communication by email from Professor Ian Lancashire of the University of Toronto.} I suspect that the number of repeated words in the EMEDD would be fewer than one percent, which would be a sizeable error of 2,000 words. Having regard for the error analysis in Tables \ref{Table 1-1} and \ref{Table 1-1Lower}, the EMEDD likely provides a reasonable estimate of the size of the lexicon at 1657.

The third source of English lexical data is the 1989 Oxford English Dictionary (OED), which has a total of 616,500 word-forms \cite{OED}. Having regard for  (\ref{eq 10}), we define $N(t_1)$, $N(t_2)$, $N(t_3)$ for the English lexicon at 1150, 1657, and 1989 as set out in Table \ref{Table 1}. Using Table \ref{Table 1}, $(\Delta t )_1 = (t_3 - t_1)$ and $(\Delta t )_2 = (t_3 - t_2)$. We solve for $r$ over the two time periods, $(\Delta t )_1$ and $(\Delta t )_2$, using (\ref{eq 10}), with the results set out in Table \ref{Table 1.1.6}.

\renewcommand{\arraystretch}{1.25}
\begin{table}[ht]
\begin{center}
	\begin{tabular}{|c|c|c|c|}\hline
	$i$ & $N(t_i)$ & Year& Lexicon Size \\ \hline
	1 & $N(t_1)$& 1150 &34,020\\\hline
	2 & $N(t_2)$&1657 & 200,000\\\hline
	3 & $N(t_3)$& 1989 & 616,500\\\hline
	\end{tabular}
	\caption{Estimated English lexicon sizes.}\label{Table 1}
	\end{center}
\end{table}

An Old English dictionary has the disadvantage and advantage of a long distance in time before the present. Since an Old English dictionary deals with words for a period before the development of the printing press, there are fewer written sources for words than for later periods of time---a disadvantage. The longer period of time from the Old English endpoint of 1150 to 1989, on the other hand, would, all other things being equal, likely result in a smaller degree of error in estimating an English lexical growth rate.

 Using the historical English dictionaries, the estimated average English lexical growth rate for the period from 1150 to 1989 is about 3.453\% per decade, and from 1657 to 1989 about 3.391\% per decade. As shown by the error calculations in Table \ref{Table 1-1}, even if one of $N(t_1)$, $N(t_2)$, $N(t_3)$ were inconsistent in counting words leading to a 10\% over count, for example, the estimate of the rate per decade for 1150 to 1989 would only be in error by about 3.33\% and for 1657 to 1989, by about 8.9\%. Both estimates, 3.391\% per decade and 3.453\% per decade, are close to the estimated average rate of increase in average IQs, which is about 3.315\% per decade. The estimated English lexical growth rate results are summarized in Table \ref{Table 1.1.6}.

\renewcommand{\arraystretch}{1.25}
\begin{table}[ht]
\begin{center}
	\begin{tabular}{|c|c|}\hline
	Time period & Average English lexical growth rate per decade\\ \hline
	1150 to 1989&3.453\%\\\hline
	1657 to 1989& 3.391\%\\\hline
	\end{tabular}
	\caption{The estimated average English lexical growth rate.}\label{Table 1.1.6}
	\end{center}
\end{table}

\subsection{Lighting efficiency}\label{Subsection lighting data}

The third set of data we consider concerns the average rate of increase in lighting efficiency. 

The lighting efficiency data is from a study by Professor Nordhaus \cite{Nordhaus1997} about the change in lighting efficiency from  1,750 B.C.E. to the year 1992. To determine the average rate of increase in lighting efficiency it is necessary to solve for $r$ using the ratio set out in  (\ref{eq 11}). The improvement in lighting efficiency  required the evaluation, by many people, of the convenience of many different lighting innovations, over a period of 3,742 years, longer than the period of time considered for English lexical growth, and much longer than the several decades of data about increasing average IQs.  Since domestic lighting is a widely used technology, involving so many  members of society, in their various roles as inventors, manufacturers, and wholesale and retail purchasers, and as users of lighting, it is likely that the problem solving required by the evaluation of lighting technology and improvements is representative of society's average general, collective problem solving capacity. Improvements in the \textit{depth} of ideas may also have facilitated improvements in the technology of lighting. 

Nordhaus studied lighting efficiency  to test whether price indexes accurately capture improvements in the material well-being of people arising from improved technologies. One of the bases of comparison he uses is what he calls the `labor price of lighting,' measured in hours of work required to pay for 1000 lumen hours, lumen being a measure of illumination. For example, Nordhaus estimated that a sesame lamp in Babylonia around 1750 B.C.E. cost about 41.5 hours of work per 1000 lumen hours, and that in 1992 a compact fluorescent light cost 0.000119 hours of work per 1000 lumen hours. If \textit{The General Collective Problem Solving Capacity Theorem} is valid, then the improvement in lighting efficiency should equal the rate of improvement in lexical efficiency, which we have estimated using English lexical growth. 

We can use (\ref{eq 11}) to find the relative increase in the efficiency of lighting from 1750 B.C.E. to 1992. It is

\begin{equation}\label{eq 24-2}
\\ \frac{ 41.5}{.000119} = 348,739.5.
\end{equation}

\noindent As Nordhaus writes, 

\begin{quote}
``an hour's work today will buy about 350,000 times as much illumination as could be bought in early Babylonia \cite[p. 33, Cowles Paper]{Nordhaus1997}.''
\end{quote}

\noindent In (\ref{eq 24-2}) we have the ratio described in (\ref{eq 11}) for use in (\ref{eq 10}), to calculate, in terms of its labor cost, the rate of increase in lighting efficiency. Based on Nordhaus's work, applying (\ref{eq 10}) for the period of $1750 + 1992 = 3742$ years, and using the result from (\ref{eq 24-2}), we find that the average rate of increase $r$ in lighting efficiency is, 

\begin{equation}\label{eq 24-4}
	 \ln \left( \frac{ 41.5}{.000119} \right) \div 3742= .00341 \ per \ year,
\end{equation}

\noindent or 3.41\% per decade.

\renewcommand{\arraystretch}{1.75}

\begin{table}[ht]
\begin{center}
\footnotesize
	\begin{tabular}{|c|c|}\hline
	Years&Average increase in lumens per hour of labor\\ \hline
	1750 B.C.E. to 1992&3.41\% per decade\\\hline
	\end{tabular}
	\caption{ The average rate of increase in lighting efficiency, based on data from \cite{Nordhaus1997}.}\label{Table 1.1.8}
	\end{center}
\end{table}

\normalsize

\subsection{Inferences and problems arising out of rate data}\label{Subsection inf and problems}

The closeness of the rate of English lexical growth, from 1150 to 1989---3.453\% per decade, which is 0.0353 high---and from 1657 to 1989---3.391\% per decade, which is 0.019 less---compared to the 3.41\% per decade increase in lighting efficiency, is remarkable. An estimate of society's average general collective problem solving rate based on a longer period of time, all other things being equal, is likely to be more accurate, as implied by Tables \ref{Table 1-1} and \ref{Table 1-1Lower}. Having regard for the error analysis summarized in Tables \ref{Table 1-1} and \ref{Table 1-1Lower}, a 10\% error in an estimate of $N(t_2)$ for a period of time from 1750 B.C.E. to 1992, 3,742 years, would result in an error in the estimate of the rate per decade of less than $1/10^{th}$ of a percent. Since the 3,742 years covered by Nordhaus's study is much longer than the periods of time considered for English lexical growth and for increasing average IQs, and since the average rate of increase in average IQs  and the average English lexical growth rate are so close to the average rate of increase in lighting efficiency, the 3.41\% per decade average increase in lighting efficiency is likely a more reliable estimate of society's general, collective, problem solving rate.  I therefore propose to use 3.41\% per decade as the estimate of society's recent average general collective problem solving rate.  If Nordhaus's lighting study is used as a benchmark, that suggests that our lexical data for 1657 is somewhat more reliable than our lexical data for 1150, which was used to calculate the average English lexical growth rate from 1150 to 1989. All four rates---the rate at which average IQs increase, the calculation of average English lexical growth for two periods of time, and the rate of increase in lighting efficiency---almost coincide.

Although the close correspondences of these calculated rates supports the hypothesis that society has a general collective problem solving rate,  these correspondences raised three problems---possible inconsistencies---for me when I first read Nordhaus's paper in the fall of 2008. 

The first problem raised by Nordhaus's lighting study was: how is it possible for the rate of problem solving to be about 3.41\% per decade back to 1,750 years B.C.E., when according to my (erroneous) calculations, done in 2005, it was 2\% per decade from 600 to 1989? I had taken 1657, the end point of the University of Toronto's Early Middle English (EMEDD) historical dictionary project, to date the size of the English lexicon in 1657, and had found an English lexical growth rate of 3.391\% per decade from 1657 to 1989. I had originally (and mistakenly) guessed that the year 600 was an appropriate dating for Old English; U of T's Old English dictionary project takes the lexicon to 1150. It seems clear in retrospect that the ending year of a dictionary should be used because the lexicon grows cumulatively. A second error of lesser import was taking the number of letters in the Old English alphabet as 24, whereas U of T's Old English dictionary project says that Old English had 22 letters. Based on the 8 letters now completed by U of T's Old English dictionary project, and assuming the number of words per letter is representative of the number of words per letter in the Old English lexicon, the estimated number of words in the Old English lexicon is 34,020  at the year 1150. These changes led to a revised English lexical growth rate, from 1150 to 1989, of 3.453\% per decade, not inconsistent with Nordhaus's results, and resolving the first problem raised by Nordhaus's paper. Ironically, the original, incorrectly observed, discrepancy in English lexical growth rates, between 2\% per decade for the years 600 to 1989 and 3.391\% per decade for 1657 to 1989, led me, in order to explain the apparent increase in the average English lexical growth rate, to an inference which likely is correct, that there are composite factors in the average English lexical growth rate, and, it follows, in the average IQ, which in turn suggests the idea of an innate IQ. 

The second problem raised by Nordhaus's lighting study was: exactly what collective problems are solved by society? While society's general, collective problem solving capacity is conceivably \textit{indirectly} tested in IQ tests, and is directly applied in solving language problems, it is more difficult to think that society's general collective problem solving capacity directly applies to \textit{individuals} inventing lighting improvements. How can the problem solving of a few inventors per generation be viewed as representative of society's \textit{general} collective problem solving capacity? This second problem raised by Nordhaus's paper study is resolved if collective problem solving is evaluative, and the resolution of this second problem raised by Nordhaus's paper has been incorporated into this article. Another possible resolution, not inconsistent with the first suggested resolution, is that the problem solving capacity of the \textit{average inventor} increases in proportion to the increase in society's general collective problem solving capacity, as the \textit{depth} of society's store of abstractions and theories increases. Inventors, like everyone else in society, benefit from the increasing depth of the abstractions they use that are a collective achievement of society.

The third problem raised by Nordhaus's lighting study concerns the factors that affect society's general collective problem solving rate. If a function of population size is a factor, how is it that the population involved in evaluating lighting innovations still exhibits the same rate of increase as exhibited by an English speaking population, despite the two populations likely being different in their sizes, individuals, and eras? Perhaps it is because the average innate, or basal, rate of problem solving is so low that a few thousand years is not enough time for differences in population to have a large effect. Perhaps for large populations, the difference in the logarithm of the population sizes is relatively small. Perhaps people are so networked that the relevant networked population is that of all of humanity.

The estimates of society's general collective problem solving rate are all consistent with the hypothesis that there is a general problem solving capacity, and are mutually confirmatory. We conclude that it is likely the case that, in recent historical times:

\begin{equation}\label{eq 25}
\begin{split}
\ \frac{d  \left|\{PS_{Society}(t)\}\right| }{dt} 
&=\frac{d  \left|Lex_{English}\right| }{dt} \\ 
&=\frac{d  \left|\{PS_{Lang}\}\right| }{dt} \\ 
&=\frac{d(IQ_{Av})}{dt} \\ 
&=\frac{d(IQ_{Society})}{dt} \\ 
&=\frac{d  \left|Lux\right|   }{dt} \\ 
&=\left(\frac{d \left| \{PS\} \right|  }{dt}\right)_{Av} \\ 
& \approx 3.41\% \ per \ decade.
\end{split}
\end{equation}

The result in  (\ref{eq 25}) compresses much compiled data---all of which required a proportionate amount of energy for researchers to compile ---into a succinct statement. Devising the concepts of IQ and intelligence testing took problem solving energy. Skilled design and administration of IQ tests to hundreds of thousands of people requires the expenditure of energy by many people. Only the problem solving work that laid the foundation for IQ testing permits researchers to have available results which permit them to estimate---spending yet more energy---the rate at which average IQ test scores increase.  

The possibility of estimating a lexical growth rate increases with the availability of historical dictionary projects.  Each word in a dictionary involves the cumulative energy of those many different people over many generations who: used the words in writing, located usages of each word, evaluated each word for its use and convenience, made a professional, learned judgment as to the appropriateness of including a word as a distinct word in the dictionary. 

Nordhaus's lighting study first required him to conceive the idea of estimating the rate of increase in lighting efficiency in order to test how well price indexes measure changes in the cost of living. In his paper, he summarizes the history of lighting, compares light outputs, finds labor costs, and calculates the increase in lighting efficiency in different ways, based on data he collected. In the appendix, he describes the steps he took to measure the light output of firewood and sesame oil lamps. 

The possibility of formulating a statement such as that in  (\ref{eq 25}) arose only recently. Nordhaus's lighting study was published in 1997. IQ testing began around 1905, Tuddenham's study was published in 1948, and much of the IQ test data used to estimate the rate of increase in average IQs was gathered since 1970. The University of Toronto's historical English dictionary projects began  around the 1990s. The accessibility of such information has been enhanced by the internet, itself the product of many solved problems.

Thus, the recent addition to society's store of solved problems has enabled an analysis that would have been impossible thirty, and perhaps even fifteen, years ago. Other possible sources of data might include the system developed by Arpad Elo used to rank chess players.\footnote{Arpad Elo's method was brought to my attention by Dr. Steve Jacobs, Toronto.}

A summary to this point is:
\begin{itemize}
	\item 	Intelligence is a problem solving capacity.
	\item 	A problem solving capacity is equivalent to a problem solving rate.
	\item 	Society's intelligence is an emergent collective phenomenon. 
	\item 	An IQ estimates an individual's problem solving rate.
	\item 	In principle, the rate of increase in the average individual problem solving rate should equal the rate of increase in society's average general problem solving rate, as in (\ref{eq Soc-IQ.30}).
	\item 	Data for three different kinds of problem solving---involving IQ tests, lexicons, and lighting---are all consistent with, and mutually confirmatory of, the existence of society having an emergent general collective problem solving capacity.
	\item 	Society's average general collective problem solving rate is close to 3.41\% per decade.
\end{itemize}

One implication of the foregoing affects a thesis, accepted by many linguists since the 1950s \cite[pp. 419--421]{OxMind}, that human facility with grammar is genetically based---that human facility with grammar is an ``innate endowment'', a``biological endowment'', or a ``language instinct.'' Society's general collective problem solving capacity applies to solving the problem of devising and improving a grammar to make the use of language more useful, convenient and efficient. Individual human beings, including young children, each with a vast network of neurons, easily learn the grammar that generations of society have invented and refined, because grammar, among other things, was designed to be easily learnable by everyone. About the proponents of grammar as a genetic endowment, Terence Deacon has written:

\begin{quote}
	``They argue that even much more extensive experience of the type that children do not get might still be insufficient to allow one to discover the abstract rules that constitute the grammar of a natural language. \ldots They assert that the source of prior support for language acquisition must originate from \textit{inside} the brain, on the unstated assumption that there is no other possible source. But there is another alternative: that the extra support for language learning is vested neither in the brain of the child nor in the brains of parents or teachers, but outside brains, in language itself''\cite[pp. 104 and 105]{Deacon1997}.
\end{quote}

\noindent From the observations to this point of the article, we argue that grammatical skill is built into language, and is the \textit{emergent} result of society's collective problem solving. It is difficult, if not impossible, for human beings to distinguish, in the exercise of their intellect, between what is innate and what is learned. The \textit{depth} of grammatical solved problems continuously and \textit{emergently} improves; human beings learn those improvements. A ``language instinct'' is an unnecessary hypothesis.  Since, moreover, it is unlikely that the brain of a human being developed a genetically endowed capacity to use grammar before the need for grammar---as a way to organize words---arose, the hypothesis of a `language' instinct is likely incorrect. Emergence is a likelier explanation. 

Another implication of the foregoing affects the puzzle about  rising average IQs. No current consensus exists on what causes increasing average IQs. The results above suggest that, while for thousands of years the physiological capacities of a human being---the capacity of a human being's brain---to solve problems has remained relatively static, \textit{the single networked mind of society has, over the same period of time, increased \emph{its} intelligence}. 

An \textit{individual's} intelligence emerges from a network of neurons. An individual's innate capacity to solve problems does not change during the individual's lifetime, but the individual's store of solved problems, as one of the factors in the composite problem solving capacity, does increase, as the individual learns more during their life. A \textit{society's} intelligence emerges from a network of brains, increasing society's intelligence not just through \textit{current} social networking, but also by society \textit{accumulating solved problems}. A society's cumulative capacity to solve problems \textit{does change} during the lifetime of the society as the society accumulates and stores more solved problems. And the lifetime of the single `mind' of society is much longer than that of any individual's mind. Language facilitates social networking, and increases society's capacity to solve problems and store their solutions. Members of society acquire most of their abstractions from society's mind.

As society improves the \textit{depth} and \textit{range} of its abstractions, average IQs increase; society's mind is the primary repository of stored intelligence---of solved problems. A rising tide (society's better abstractions) raises all boats (average individual intelligence). With better conceptual tools, a person (and an inventor) can think more efficiently, applying an unchanging innate problem solving capacity. 

Members of society, through social networking, obtain solutions to problems solved by the expenditure of \textit{other people's energy}. The advantage of learning solved problems---acquiring knowledge---is that \textit{the energy cost of learning existing solutions} is much less than \textit{the energy cost of solving  problems from scratch}. 

In 1832, John Austin observed:

\begin{quotation}
	``If our experience and observation of particulars were not \textit{generalized}, our experience and observation of particulars would seldom avail us in \textit{practice}. To review on the spur of the occasion a host of particulars, and to obtain from those particulars a conclusion applicable to the case, were a process too slow and uncertain to meet the exigencies of our lives. The inferences suggested to our minds by repeated experience and observation are, therefore, drawn into \textit{principles}, or compressed into \textit{maxims}. These we carry about us ready for use, and apply to individual cases promptly or without hesitation: without reverting to the process by which they were obtained; or without recalling, and arraying before our minds, the numerous and intricate considerations of which they are handy abridgments'' \cite[p 49]{Austin1832}.
\end{quotation}

Indirectly, knowledge increases the effective problem solving output of an individual---as if the individual had acquired increased energy resources.  If we consider \textit{problem solving as a technology}, then the increase in the number of solved problems over time is consistent with Romer's thesis \cite{Romer1990} that \textit{endogenous technological growth} promotes economic growth: \textit{endogenous problem solving} promotes economic growth.  

An individual's problem solving capacity may seem to originate with the individual. But an individual's innate IQ is multiplied by a logarithmic function of the number of problems solved by society that the individual \textit{has learned}. The \textit{multiplier} is to be credited mostly to society, not to the individual who learns society's solved problems. It is  difficult---if not impossible---for an individual to distinguish their own innate problem solving capacity from  the benefits of having had access to society's store of solved problems (including the benefits of society's infrastructure---society's solved problems made physically manifest). The philosopher Paul Feyerabend seems to have been alluding to a similar observation when he remarked that

\begin{quotation}\label{Feyerquote}
	``We may, of course, abstractly subdivide this process into parts, and may also try to create a situation where statement and phenomenon seem to be psychologically apart and waiting to be related. (This is rather difficult to achieve and is perhaps entirely impossible.) But under normal circumstances such a division does not occur; describing a familiar situation is, for the speaker, an event in which statement and phenomenon are firmly glued together.
	
	This unity is the result of a process of learning that starts in one's childhood. From our very early days we learn to react to situations with the appropriate responses, linguistic or otherwise. The teaching procedures both \emph{shape} the `appearance', or `phenomenon', and establish a firm \emph{connection} with words, so that finally the phenomena seem to speak for themselves without outside help or extraneous knowledge. They \emph{are} what the associated statements assert them to be. The language they `speak' is, of course, influenced by the beliefs of earlier generations which have been held for so long that they no longer appear as separate principles, but enter the terms of everyday discourse, and after the prescribed training, seem to emerge from the things themselves\ldots'' \cite[p. 57]{Feyerabend1993}.
\end{quotation}

One problem raised by the preceding discussion is: \textit{what is the innate, or basal, problem solving capacity of an individual}?

\section{Calculating the benefit of networking}\label{Section Networking benefit}

\subsection{Preliminary considerations}\label{Prelims}

Since the average rate of increase in individual IQs equals the average rate of increase in society's IQ, as expressed in (\ref{eq Soc-IQ.40}), an increase in an individual's problem solving capacity must be due to networking.  Neurons in a single brain can solve problems. More neurons, in more brains, can solve more problems. A society has  more neurons than any individual member of society. Therefore, all other things being equal, an increase in a society's population size should increase the society's problem solving capacity. 

Let $Pop (t)$ represent population size at year $t$.  Then the immediately previous  observation implies, consistent with (\ref{eq Networking-10.1000}), that

\begin{equation}\label{eq 26}
	 \frac{d |\{PS_{Society}(t)\}|}{dt}\propto f(Pop(t)).
\end{equation}

\noindent If the left side in (\ref{eq 26}) is society's \textit{average} general collective problem solving rate, then the value of the function $f(Pop(t))$ on the right side should be an average applicable for the same time period. The economist M. Kremer observes that 

\begin{quote}
``the growth rate of technology is proportional to total population'' \cite[p. 681]{Kremer1993}.
\end{quote} 

\noindent Consistent with \textit{The General Collective Problem Solving Capacity Hypothesis}, we regard technological innovation as a particular result of the application of society's general collective problem solving capacity. We propose that society's general collective problem solving \textit{rate} is proportional to some \textit{function of the total population}, $f(Pop(t))$.

Suppose that the population is unchanging, and consider the effect of a store of solved problems on society's general collective problem solving rate. When an individual is confronted with a problem, the individual can use society's existing store of solved problems as a resource; the individual can vary an existing solution, blend together solutions, and can create a new solution to an existing or new problem. New problem solving can use procedures---methods that are themselves solutions to the problem of \textit{how to solve} other problems---that have been used in the past to solve problems. The larger the existing store of solved problems is, the more ways there are to apply an individual's innate, or basal, problem solving capacity to a given problem: there are more ways to tackle the problem. Accordingly, society's general collective problem solving rate should be greater when the store of solved problems is greater. Thus, consistent with (\ref{eq eta-10.50}),

\begin{equation}\label{eq 27}
	 \frac{d[|\{PS_{Society} (t)\}|]}{dt} \propto g(|\{ PS_{Society}(t)\}|).
\end{equation}

\noindent If the left side in (\ref{eq 27}) is society's \textit{average} general collective problem solving rate, then the value of the function $g(|\{ PS_{Society}(t)\}|)$ on the right side should also be an average applicable for the same time period. Kremer writes, consistent with (\ref{eq 27}), that research productivity depends on the existing level of technology \cite[p. 689]{Kremer1993}---what we would characterize as an existing store of solved problems. (Kremer refers to a then unpublished article by Charles Jones, now \cite{Jones1995}.)

A function $h$, assumed to be independent of the functions $f(Pop)$ and $g( |\{PS_{Society}(t)\}|)$, represents the effect on society's general problem solving capacity due to the society's physical environment, climate, and infrastructure, to the extent that their contributions to problem solving capacities are not already reflected in $f$ in (\ref{eq 26}) and $g$ in (\ref{eq 27}). Even though a society's \textit{current} infrastructure is due to the problem solving capacities of its population in the past, and to its store of already solved problems, the inherited infrastructure is not created by society's \textit{current} problem solving capacity, and may be considered to be independent of the society's \textit{current} problem solving capacity. We may suppose that, at a time $t$, 

\begin{equation}\label{eq 28}
\\\frac{d|\{PS_{Society}(t)\}|}{dt} \propto h( |\{Infrastructure_{Society}(t)\}|).
\end{equation}

The function $h$ will play a possible role later in this article. For now, the analysis is simpler without it. 

From the relationship between society's general collective problem solving capacity and the functions $f$ and $g$ expressed in (\ref{eq 26}) and (\ref{eq 27}), we infer that 

\begin{equation}\label{eq 29}
\\ \frac{d |\{PS_{Society}(t)\}|}{dt} \propto f( Pop(t))\times g( |\{PS_{Society}(t)\}|),
\end{equation}

\noindent a result similar to (\ref{eq Networking-10.1000}).

Earlier in this article we inferred that, since the physiology of the human brain has been relatively unchanged these past several thousands of years, and the problem solving capacity of an individual human brain is a function of the physiological capacities of that brain, the average innate problem solving capacity has also remained unchanged during that period of time. Since a problem solving capacity is equivalent to the problem solving rate, it follows that
\begin{equation}\label{eq 32}
	 IQ_{Innate} = \left(\frac{d |\{PS_{Average \ Innate}\}|}{dt}\right)_{Av} = m,
\end{equation}

\noindent where $m$ is a constant. (\ref{eq 29}), together with (\ref{eq 32}), implies that

\begin{equation}\label{eq 33.10}
	 \left(\frac{d |\{PS_{ Individual(t)}\}|}{dt}\right)_{Av} \propto m \times \frac{d |\{PS_{Society}(t)\}|}{dt}, 
\end{equation}

\noindent and so, since $m$ is a constant,

\begin{equation}\label{eq 33.20}
	 \left(\frac{d |\{PS_{Individual(t)}\}|}{dt}\right)_{Av} \propto \frac{d |\{PS_{Society}(t)\}|}{dt}.
\end{equation}

\noindent In light of the \textit{statistical evidence} that society has a general collective problem solving rate, we infer that 

\begin{equation}\label{eq 34}
	\left(\frac{d |\{PS_{Individual}(t)\}|}{dt}\right)_{Av} 
= \frac{d |\{PS_{Society}(t)\}|}{dt}
\end{equation}

\noindent consistent with \textit{The General Collective Problem Solving Capacity Hypothesis} and (\ref{eq Soc-IQ.40}).

Since society has increased its store of solved problems over the generations, the average problem solving capacity of a human being, with access to an increasing number of society's solved problems, has also increased, in proportion, over those generations. If we extrapolate back in time, the reverse observation would appear to apply. The farther back in time one goes, the smaller is the value of the function of the number of society's solved problems that multiplies an individual's innate problem solving capacity. Far back in time, the difference between the problem solving capacity of an individual human being and, for example, other mammals or primates, would be considerably smaller than present day differences. 

From  (\ref{eq 25}),

\begin{equation}\label{eq 35}
	 \frac{d |\{PS_{Society}(t)\}|}{dt} \approx 3.41\% \ per \ decade,
\end{equation}

\noindent society's general collective problem solving rate. Now applying  (\ref{eq 29}) and (\ref{eq 32}), we have

\begin{equation}\label{eq 36}
\begin{split}
	 & \left(\frac{d |\{PS_{Innate}(t)\}|}{dt}\right)_{Av} \times f([ Pop(t)])\times g( |\{ PS_{Society}(t) \}| )\\
&=\frac{d |\{PS_{Society}(t)\}|}{dt} \\
&=3.41\% \ per \ decade.
\end{split}
\end{equation}

If we can find functions $f$ and $g$ for $f( Pop(t))$ and $g( |\{ PS_{Society}(t) \}| )$, and quantify their values for a given society at a given time, then we will be able to quantify the current average innate, or basal, problem solving \textit{capacity} of an individual human being---or, equivalently, the average innate, or basal, problem solving \textit{rate}---the problem solving capacity that would exist in the absence of social contact and in the absence of an inherited body of knowledge.

What functions might $f$ and $g$ be? In subsection \ref{Subsection PS as evaluation}, we inferred, based on an analogy between absolute temperature and IQ as metrics, that $f=g=\log$. We now explore the nature of $f$ and $g$ from a different perspective, to find the parameters $C$ (the proportion multiplier) and $S$ (the base of the logarithmic function) for a society's store of solved problems and for its population. 

Consider societies and lexicons as hierarchically structured networks. People network with other people. Words network with other words; ideas network with other ideas. The philosopher David Hume wrote:

\begin{quotation}
	``As all simple ideas may be separated by the imagination, and may be united again in what form it pleases, nothing would be more unaccountable than the operations of that faculty, were it not guided by some universal principles, which render it, in some measure, uniform with itself in all times and places. Were ideas entirely loose and unconnected, chance alone would join them; and it is impossible the same simple ideas should fall regularly into complex ones (as they Commonly do) without some bond of union among them, some associating quality, by which one idea naturally introduces another. This uniting principle among ideas is not to be considered as an inseparable connexion; for that has been already excluded from the imagination: Nor yet are we to conclude, that without it the mind cannot join two ideas; for nothing is more free than that faculty: but we are only to regard it as a gentle force, which commonly prevails, and is the cause why, among other things, languages so nearly correspond to each other; nature in a manner pointing out to every one those simple ideas, which are most proper to be united in a complex one'' \cite[Book I, Sec. IV]{Hume1739}. 
\end{quotation}

Networks use energy; networks \textit{must} be \textit{structured} to use energy efficiently in order to successfully compete for their environment's resources, and to thrive. If there is a general efficiency standard---`a uniting principle' based on an \textit{emergent} `gentle force'---for  structuring a network regardless of what the constituent nodes are, then an efficiently structured network should conform to that general efficiency standard. We might think of the general efficiency  standard as an ideally efficient network, or, an \textit{Ideal Network}. We infer that if networks are structured to conform as closely as possible to  an Ideal Network, then networks should have analogous structures. Since communication is vital to the survival of human society, we should expect that both social networks and lexical networks have adapted to conform as closely as possible to  an Ideal Network; this is the `uniting principle.' If that is so, the same function should apply to describe the benefit of networking in each. We infer, on that basis as well, that in  (\ref{eq 29}) 

\begin{equation}\label{eq 37}
\\ f = g.
\end{equation}

There is another argument that suggests $f$ = $g$, besides the argument that an Ideal Network exists, to which energy efficient networks conform. Suppose a network consists of $n$ ideas, abstractions, or solved problems. We can infer that finding each of the functions $f$ and $g$ is equivalent to finding the answer to the following: find all possible ways that a number of information units of size $r$ can be distributed among $n$ nodes. The distribution of $r$ information units among $n$ receivers is similar to a problem that physicists, including James Clerk Maxwell, Ludwig Boltzmann and J. Willard Gibbs, worked on from about 1870 to 1900, now considered to be part of statistical mechanics.  Their concern was not with information units but rather with energy units. But, mathematically, the two distribution problems are equivalent. Boltzmann followed a procedure equivalent to using the factorial expression

\begin{equation}\label{eq 38}
	 \frac{(n+r-1)!}{(n-1)!r!} 
\end{equation}

\noindent to calculate the number of different ways of distributing $r$ energy units among $r$ receivers. When $n$ is large, the value of (\ref{eq 38}) is estimated using Stirling's approximation.  Stirling's approximation is used to calculate the value of a factorial that involves large numbers.  One conclusion of statistical mechanics is that, assuming the independence of the velocity of the energy units being distributed, the most likely distribution is the one in which the energy is equally distributed to space elements of an equal size. This isotropic distribution is, in a sense, the most efficient distribution of energy units to a given volume of space.

The number of combinations of energy units possible is highest when the distribution to all equal energy receivers is equal, that is, when the probability of a receiver receiving an energy unit is equal for all receivers---when all receivers have an equal capacity for receiving energy. The equal distribution of energy units is the statistically most likely distribution. One might say that when the number of ways, or paths, in which to create an energy distribution is maximized, the number of ways for energy to be distributed to the recipients is maximized; this occurs when the probability of energy being distributed to a recipient is equalized. 

We can consider a network with $n$ nodes to be analogous to a statistical mechanical system of $n$ elements. We can consider problem solvers, or solved problems, such as words, to be nodes in a network. All the energy provided to problem solvers for problem solving, would, if the problem solvers were perfectly efficient, be entirely used to produce solved problems. Since a solved problem is energy-equivalent to the energy input used to solve it---energy in equals energy-equivalent information out---the network of problem solvers should be isotropic with respect to energy use if the energy input is isotropic. Since the analysis of a network of problem solvers and a network of solved problems can both be analogized to the statistical mechanical problem of distributing energy (or information) among $n$ receivers, there is further support for the idea that $f$ = $g$. Network energy distribution systems are structurally analogous.

The question, how much does networking increase the innate, or basal, problem solving capacity of an individual, is analogous to a question about how to maximize the efficiency of the distribution of energy in statistical mechanics. Because of the fundamental role of energy in all kinds of systems, it is likely not a mere coincidence. 

\subsection{A network capacity multiplier, based on the network's path length}\label{Subsection, capacity multiplier}

To assist us in deriving the function $f$ used in $f(Pop)(t)$, let's consider a social network in which information is distributed by an individual to one other individual at a time. An individual $A$ is \textit{one step} away from individual $B$ if $B$ directly transmits information to $A$. $A$ is \textit{two steps} away from individual $C$ if $C$ first transmits information to $B$ who then transmits $C$'s information to $A$. And so on. There exists some distance---a \textit{pair-wise, or bi-nodal, distance} expressible in steps---between any two individuals in the network.

Suppose there exists a network of finitely many individuals $n$, any two of which are separated by a finite distance, measured in \textit{steps}.  To simplify things, assume that all transmitters transmit  information at the same rate, and that all receivers receive  information at that same rate.  These assumptions result in the network being \textit{isotropic}, both in the transmission and in the reception of information. The total of all the \textit{pair-wise distances}, \textit{in steps}, is finite, because every pair-wise distance is finite, and for the network there are only a finite number of all possible pairs. If we divide the total of all the pair-wise distances for all distinct pairs by the total number of possible pairs, we obtain a finite \textit{average} distance, $S$, measured \textit{in steps}, and which in network science is given the name: the \textit{path length}. That is, the average pair of nodes, or individuals, in the network connects in an average of $S$ steps. Colloquially, the path length is called \textit{degrees of separation}. Just as absolute temperature measures the quantity of heat in a system, and IQ measures  an individual's intelligence, a `step' is a metric that measures a distance. Here, the \textit{average distance} between nodes in a network is measured by the average number of steps from one to the other. In a sense, the path length is a more fundamental metric than absolute temperature (where the degree measure has reference to the freezing and boiling point of water); at any given point of time, a network's path length is intrinsic to that network. For the path length, no reference scale is required other than the network itself.

There exists some number $\eta$ (the small Greek letter, eta, $\eta$) such that

\begin{equation}\label{eq 39}
\\ S^{\eta}= n,
\end{equation}

\noindent for a network consisting of $n$ nodes, where $S$ is the network's path length.

Along a path $S$ steps long there are located $S$ possible information sources. Notionally only, a person has the \textit{capacity} to receive information from the whole network in the following way. We consider a source that is $S$ steps away \textit{the first cluster generation}, a source $2 \times S$ steps away \textit{a second cluster generation}, and in general a source $\eta \times S$ steps away a $\eta $th \textit{cluster generation}. (\ref{eq 39}) shows that there are $\eta$ cluster generations. Since \textit{all} nodes in the network are only $S$ steps apart, \textit{all} the network's nodes are in \textit{each} cluster generation. Looking at cluster generations collectively, since every node in the network can be connected to every other node in an average of $S$ steps, it does not matter in what cluster generation a receiver resides; the transmitter can reach it in an average of $S$ steps. 

The structure of cluster generations---the scaling 

\begin{equation}\label{eq scaling10.10}
 \overbrace{S \times S \times S \times \dots \times S}^{\eta \ S's}
\end{equation}

\noindent implicit in (\ref{eq 39})---is such that when the exponent of $S$ goes up by one, something in the network must be scaling up by a factor $S$, while leaving the number of nodes per cluster generation unchanged. We infer that the number of clusters per cluster generation scales up by $S$ while the number of nodes per cluster scales down by $1/S$; in that way the number of nodes per cluster generation is constant. It must be that cluster generations with more clusters are nested in cluster generations with fewer clusters, because  \textit{each cluster generation} contains \textit{all} the network's nodes. We denote each generation by the term \textit{cluster generation} because while all the network's nodes are in each notional cluster generation, each distinct cluster generation varies in the number of clusters and the number of nodes per cluster, by some power of $S$.

\paragraph{Isolated clusters.} Now consider any \textit{single cluster} of $S$ nodes. Suppose the pair-wise, bi-nodal rate of information transmission in that cluster is $S$ steps per second, and suppose that applies to any pair of nodes. Then the average time for a node in the network to connect to any other node \textit{within that single cluster} (which in turn is in a single cluster generation) is one second, because the average bi-nodal distance is $S$ steps. To traverse $\eta$ \textit{disconnected} cluster generations---that is, $\eta$ distinct networks each of $n$ nodes---would  take $\eta$ seconds at the pair-wise, or bi-nodal, rate of $S$ steps per second. 

\paragraph{Networked clusters.} \textit{For networked nodes}, $S$ steps is the average pair-wise distance for \textit{all possible nodal pairs} in the network; for one node to connect to any other node in the same network at a pair-wise rate of $S$ steps per second gives the node a \textit{capacity} to traverse all $\eta$ cluster generations in only $S$ steps, or, in only one second---not $\eta $ seconds---because all nodes in the network are (an average of) only $S$ steps apart. The capacity to traverse $\eta$ separate clusters in $\eta$ \textit{separated} cluster generations at the rate of $S$ steps per second is equivalent to having the capacity to traverse all \textit{networked} $\eta$ cluster generations at the pair-wise rate of $S$ steps per second. Thus, upon networking, an isolated \textit{pair-wise or bi-nodal rate} of $S$ steps per second becomes \textit{a Network Rate} equivalent to $\eta S$ steps per second. 

\hfill\\
Let's try another way of looking at this. 

Let $\left\|a_{i}, b_{i}\right\|_{i}$ be the distance in steps between 2 nodes, $a_i$ and $b_i$, in the $i$th cluster generation of a network, $N$, with $n$ nodes.

Let $S$, as above, be $N$'s path length, the average distance in steps between any pair of nodes in $N$. From (\ref{eq 39}),

\begin{equation}\label{eq Multiplier.10.10}
\\ \log_{S} ( n )= \eta.
\end{equation}

Now, from (\ref{eq 39}), within the $i^{th}$ cluster generation,

\begin{equation}\label{eq Multiplier.10.20}
	 \left\|a_{i}, b_{i}\right\|_{i} = S, \forall a_{i}, b_{i} \in N, a_{i} \neq b_{i}, \forall i \leq \eta. 
\end{equation}

What the discussion about pair-wise distances implies is that the same energy used for a transmission of $S$ steps between any two nodes in $N$, also allows a transmission that traverses the equivalent of all $\eta$ cluster generations, which in steps would be:

\begin{equation}\label{eq Multiplier.10.30}
\begin{split}
	 & \left\|a_{1}, b_{1}\right\|_{1}+\left\|a_{2}, b_{2}\right\|_{2} + \ldots +\left\|a_{\eta}, b_{\eta}\right\|_{\eta} \\
& = \left\|a_{i}, b_{j}\right\|, \forall a_{i}, b_{j} \in N, a_{i} \neq b_{j}, \forall i, j \leq \eta \\
& = \eta \times S, \forall a, b \in N, 
\end{split}
\end{equation}

\noindent or, alternatively, 

\begin{equation}\label{eq Multiplier.10.40}
\begin{split}
	 & \left\|a_{1}, a_{2}\right\|_{1}+\left\|a_{2}, a_{3}\right\|_{2} + \ldots +\left\|a_{\eta -1}, a_{\eta}\right\|_{\eta} \\
& = \left\|a_{i}, a_{j}\right\|, \forall a_{i}, a_{j} \in N, a_{i} \neq a_{j}, \forall i, j \leq \eta \\
& = \eta \times S. 
\end{split}
\end{equation}

\noindent $S$ `spans' the first lines  in (\ref{eq Multiplier.10.30}) and (\ref{eq Multiplier.10.40}), because the average distance between any two nodes in $N$ is $S$; $S$ steps traverses all $\eta$ cluster generations. This aspect of a network described by (\ref{eq Multiplier.10.30}) shows part of the significance of the path length as a network parameter. 

The \textit{same capacity} that enables a signal to traverse a distance of $S$ steps in an \textit{isolated} 2 node system, enables the same signal to traverse the equivalent of $\eta \times S$ steps in a fully \textit{networked} system.\footnote{An example of Nature's economy, subtlety and wit.}

The foregoing discussion leads to the following:

\begin{Isotropic Network Rate Theorem}
If non-networked nodes in an isotropic information network of $n$ nodes transmit information pair-wise at the bi-nodal rate $r$, then, when the nodes are networked, the \textit{network information transmission rate} is equivalent to  $\eta r$, where
\begin{center}
$\log_{S} \left( n \right)= \eta$. \\
\end{center}
\noindent The same energy that permits an isolated bi-nodal (2-node) network to transmit at the rate $r$, results in a network rate of transmission at the rate $\eta r$.  Or in other words, the same energy proportional to $S$ that spans an isolated bi-nodal (2-node) network, spans all possible bi-nodal pairs when they are isotropically networked. 
\end{Isotropic Network Rate Theorem}

Since not all networks are isotropic---people in a network do not have uniformly equal relationships to all other people in the network---\textit{The Isotropic Network Rate Theorem} does not describe the most general network structure. Before generalizing the theorem to the non-isotropic situation, we discuss why \textit{The Isotropic Network Rate Theorem} is, implicitly, a statement about the thermodynamics of a network. 

Suppose it takes the exact same amount of energy, $\epsilon$, to transmit a unit of information one step. Then it takes $S$ energy units $\epsilon$, that is $S\epsilon $ energy units, to transmit one information unit $S$ steps. If $S$ scales $n$, as (\ref{eq 39}) suggests, then it is equally the case that $S\epsilon$ energy units  scales $(S\epsilon)^{\eta}=n$ energy units. That is, if
\begin{equation}\label{Eq logEquiv10.20}
	S^{\eta} = n
\end{equation}
\noindent implies
\begin{equation}\label{eq 41}
\begin{split}
\log_{S} ( n )&=\log_{S} ( S^{\eta} )\\
 &= \eta,
\end{split}
\end{equation}
\noindent it must be that
\begin{equation}\label{Eq logEquiv10.40}
	(S \epsilon)^{\eta} = n
\end{equation}
\noindent implies
\begin{equation}\label{eq 41-2}
\begin{split}
\log_{S \epsilon} ( n )&=\log_{S \epsilon} ([S \epsilon] ^{\eta})\\
&= \eta .
\end{split}
\end{equation}

The path length appears to scale a network of $n$ nodes, but what is really occurring is that the network's energy resources are being scaled by an amount of energy proportional to $S \epsilon$. 

Energy scaling, which is due to nodes networking, confers an enormous energy advantage: two nodes, \textit{isolated} from any other nodes, that have the capacity to transmit information to each other at a rate of $S$ steps per time unit, have their capacity to transmit information boosted to a rate of $\eta S$ steps per time unit \textit{when networked}. Because of this multiplicative effect of networking on the pair-wise, or bi-nodal, transmission rate for two isolated nodes, an isotropic network allows the  parts of a network to communicate faster than it is possible for any two individual nodes to communicate outside of the network. In view of this multiplicative effect, we may suppose that, for nodes with the same bi-nodal capacities, no isolated bi-nodal transmission of information can be faster than the network transmission rate within an isotropic network. This inference may be styled

\begin{The Ideal Network Theorem}
For a set of nodes with the same bi-nodal capacity to transmit information, no isolated bi-nodal transmission of information can be faster than the network transmission rate within an isotropic network. 
\end{The Ideal Network Theorem}

The scaling intrinsic to an \textit{isotropic} network results in an optimal rate of information transmission. An optimal rate of information transmission suggests that the system, in this case a network, is using energy as efficiently as possible. The isotropic network is therefore an Ideal Network. This is additional support for the view suggested earlier in this article that there is a general efficiency standard, an Ideal Network, for structuring a network regardless of what the constituent nodes are.

\subsection{Deriving the General Network Rate Theorem}\label{Subsection, Net Rate Thm}

Actual social and other networks, however, are not \textit{perfectly} isotropic. In social and other networks, not all adjacent persons and nodes actually connect equally, but only some of them. For any given node, only some proportion of its adjacent nodes are connected to it. The average of all these proportions for all the nodes  in a network is called the clustering coefficient, usually denoted by the letter $C$ in the science of networks. Measurement of $C$ helps reveal the role of hubs in networks, among other things \cite{WS1998}.

With respect to information transmission in a network, since the network's nodes on average connect (locally) to a proportion $C$ of their adjacent nodes, only a proportion $C$ of all the network's possible paths are actually (globally) available for information transmission. (Global and local efficiency is considered in \cite{Latora2001}.) In an isotropic network, all paths are available and $C = 1$. 

If $0 < C < 1$, then the available energy used to transmit information traverses only a proportion $C$ of the network's possible paths. The energy efficiency of information transmission is proportionately reduced; each cluster generation only receives a proportion $C$ of the energy that would be otherwise available if all possible paths existed. Hence the multiplicative effect of networking is then only

\begin{equation}\label{eq 42}
	 C \log_{S} \left( n \right)= \eta.
\end{equation}

We  propose, the

\begin{General Network Rate Theorem} Suppose all two node networks in a network of $n$ nodes have a \textit{bi-nodal} (energy or information) transmission rate of $r$. When networked, the nodes have a \textit{Network Transmission Rate} that is $r \times C \log_{S} (n)$, where $C$ is the clustering coefficient for the network, and $S$ is its path length.
\end{General Network Rate Theorem}

The formula in (\ref{eq 42}) is the same as the formula found in subsection \ref{Subsection PS-av=PS-soc}, such as in (\ref{eq eta-10.60}); a proportionality factor $C$ and a logarithmic function are factors in individual intelligence. Thus we have an answer to the questions posed in subsection \ref{Subsection PS-av=PS-soc} about what the parameters $C$ and $S$ represent. The parameters are those used in network science: $C$ is the clustering coefficient, and $S$ is the path length, of a network.

In the discussion about the problem solving capacity of individuals in a society, we noted that the individual's innate, or basal, rate is multiplied, through social networking, by society's problem solving capacity. From the foregoing discussion, we see also that a pair-wise, or bi-nodal, transmission rate is multiplied by networking. For the same network, both the innate, or basal rate, and the pair-wise, or bi-nodal, rate have identical networking attributes. We therefore have: 

\begin{The Equivalency of the Innate and Bi-nodal Rates Theorem} The innate, or basal, capacity of an individual to transmit or receive information---solved problems--- is the same as the individual's isolated pair-wise, or bi-nodal, capacity to transmit information and to receive information. 
\end{The Equivalency of the Innate and Bi-nodal Rates Theorem}

The physiologically innate, or basal, problem solving capacity of an individual can not change due to networking; the innate capacity is fixed. But networking increases the individual's problem solving capacity, as implied by (\ref{eq Networking-10.1000}). The individual, when networked, has different and greater capacities.  How is it possible for the innate, or basal, capacity to be fixed, and the pair-wise, or bi-nodal, rate to be fixed, and yet the networked capacity of an individual to be greater? 

When the innate, or basal, problem solving capacity is applied within an isolated bi-nodal network, the bi-nodal problem solving rate and the network's problem solving rate \textit{are the same}. When the innate, or basal, problem solving capacity is applied within a network of $n$ nodes, $n> S$, the bi-nodal rate and the network rate are \textit{not the same}; the network's rate is $ C \log_{S} (n)=\eta$  times larger. Since the innate, or basal, problem solving capacity is fixed, it must be that the higher network problem solving capacity that results from networking is due to this: the innate, or basal, problem solving capacity has \textit{more ways to be applied}. 

The number of ways in which the innate, or basal, problem solving capacity can be applied is increased by $ C \log_{S} (n)=\eta$ times. In a sense, the larger the $\eta$ of an information network is, the more degrees of freedom there are in the exercise of an innate problem solving capacity. Each scaling of the lexicon by its path length, each increase in the exponent of a network's scaling factor $S$ from $k$ to $k+1$, represents another scaled path along which the problem solving capacity can go, at the rate of $S$ steps per choice of path. The same observation applies to any network of solved problems. It also applies to transmission of information within a social network. For each step in a path length of $S$ steps, a socially networked person has $S$ path choices. Each scaling of the population network by its path length represents another path along which information can be transmitted, at the rate of $S$ steps per choice of path.

\textit{The Equivalency of the Innate and Bi-nodal Rates Theorem} suggests that individuals do not increase their physiologically innate, or basal, problem solving capacity by networking; instead, networking just increases the number of paths along which their innate, or basal, problem solving capacity can be exercised. All other things being equal, a person \textit{A} biking in a town with 100 streets has not seen more town streets---paths---than a person biking in a town with one street---path---because \textit{A} is innately more adventurous or intelligent. It's just that \textit{A} has more choices of places to go. The person who has a one path bi-nodal capacity can use it on \textit{any} number of paths, \textit{given the opportunity}. An individual might, due to the subtle way a network multiplies individual capacity, fail to appreciate the contribution of fortuitous circumstances to their individual capacity, as suggested in the Feyerabend quote  at page \pageref{Feyerquote}.

Ants of the same ant colony travel diverse paths in their quest for a food source. There is an element of randomness in an individual ant's search for food. But collectively, ants explore all their available options, and that multiplies their general collective problem solving capacity. When an ant, or a group of ants, discovers a path to a food source that is superior to the alternatives, many more ants will follow that path. The other ants favor those paths, out of those many tested, which have proven to be the most advantageous. Similarly, human societies, through their individual members, explore diverse solutions to general collective problems. When solutions are found that society collectively evaluates as being superior to the alternatives, many more people will adopt those solutions, so that those solutions are preserved in society's store of solved problems. 

Another way of looking at it is this. Mathematically, it appears that an individual \textit{A}'s physiologically innate problem solving rate $m$ is multiplied by $\eta$ when  \textit{A} has a set of knowledge, represented by $\left\{PS\right\}_{A}$. Here, $\left\{PS\right\}_{A}$ is what \textit{A} knows---the set of solved problems that \textit{A} knows. So \textit{A's} average problem solving capacity is $(\eta \times m) \times \left|\left\{PS\right\}_{A}\right|$. What the hypothesis above says is that this is mathematically correct, but conceptually wrong. The number of solved problems produced by applying $m$ is $m \times \left(\eta \times \left|\left\{PS_{A}\right\}\right|\right) $; the innate, or basal, problem solving capacity, equivalent to the problem solving rate $m$, has $\eta$ times as many paths---$\eta \times |\{PS\}|$--- it can take. A person may perceive that their innate problem solving capacity is $m \times \eta $, but $m$ has not altered; after an individual is born, what changes the individual IQ is the number of solved problems that the individual has learned. Intelligence is the capacity to solve problems. In a society, an individual's problem solving capacity is the product of the factors set out in (\ref{eq Networking-10.1000}). Innate capacity alone does not determine individual problem solving capacity.

This discussion suggests a 

\begin{Capacity Multiplier Theorem} \label{CapMultThm} The physiologically innate problem solving capacity of an individual is multiplied by the number of ways in which that capacity can be exercised when it is applied to a network of $n$ nodes. The multiplicative factor that multiplies an individual's problem solving capacity is $ C \log_{S} [N(t)] $, where $N(t)$ enumerates the amount of information available to the individual at a time $t$.
\end{Capacity Multiplier Theorem}

\subsection{Isotropy and the natural logarithm}\label{Subsection, Nat log}

If isotropy underlies, or leads to, the maximum possible rate of information transmission in a network, perhaps isotropy also, by analogy, explains the maximum rate of energy transmission in our universe. According to the big bang theory of cosmological origins, the universe began as a burst of radiating energy from a point source. In the 1960s, scientists found a uniform cosmic background radiation in the sky, in all directions, at a temperature consistent with the big bang theory. In the 1990s, astronomical observations by a device called the Far-Infrared Absolute Spectrophotometer, housed in a satellite called the Cosmic Background Explorer (\textit{COBE}), measured the spectrum of the cosmic microwave background radiation as being isotropic to within one part in 100,000 \cite{Fixsen1996}. The universe is highly isotropic.

Consider a creation scenario for a universe. Suppose that before a universe begins, at what might be called a singularity---a point of no dimensions existing outside of and before time, and outside of a universe that does not yet exist---there is an enormous amount of energy, $\omega$. At creation, the energy isotropically fractures into $S$ \textit{energy clusters} each with $\omega/S$ \textit{energy units}. Assume, for simplicity's sake, that $S$ is unchanging over time, and so acts as a consistent scaling factor. Suppose that each of the first generation energy clusters  fractures at the rate $S$ and that process continues. After $k$ time units, there are $S^{k}$ energy clusters, each with $\omega/S^{k}$ energy units, and the process continues. The universe's energy cluster generations are all nested in a hierarchy: sub-atomic particles, atoms, molecules, stars, galaxies, clusters of galaxies. The volume of a $k$th level cluster is $V_{k}$. The energy density of a $k$th level cluster with volume $ V_{k}$, within a $k$th level cluster generation, is

\begin{equation}\label{eq 42-01}
	 \frac{\omega/S^{k}}{ V_{k}}.
\end{equation}

Since there are $S^{k}$ energy clusters in each cluster generation, the amount of energy in a cluster generation is calculated as follows

\begin{equation}\label{eq 42-03}
\\ \left( \frac{\omega/S^{k}}{ V_{k}} \right) \times (S^{k} V_{k})\\= \omega.
\end{equation}

In other words, the total energy of the expanding universe remains constant, though the  energy density decreases as the number of cluster generations increases. In such a universe, there would be no center. All parts of the universe would appear to expanding from any vantage point. 

If we consider the creation scenario for \textit{a universe}---not necessarily \textit{our} universe---as the target of the following analogy and a social network as its source, the following correspondences apply: the path length $S$ in a social network corresponds to $S$ as a scaling factor in the universe; a $k$th cluster generation, to use the terminology earlier in this article, appears in both the source and target of the analogy; and nodes per cluster generation in a social network correspond to energy units per cluster generation in the universe.

Since the two systems, a social network and the universe described above, have analogous scaling features, as a  heuristic we may infer that perhaps the universe began with the \textit{isotropic} distribution of energy \textit{because} that allowed for the fastest possible transmission of energy between parts of the universe, just as, for a given amount of information, the isotropic distribution of information  allows for the fastest transmission of information in an information network. 

If our universe began in a way similar to the theoretical universe above, then a social network should not be the source of an analogy that models the universe, but rather the target. The universe preceded social networks; the universe is the \textit{source} of the analogy. 

More likely, the structure of a social network is constrained by the isotropic way in which our universe began; the manner of creation of the universe governs the thermodynamics of social networks. In the energy scaling model of a universe, the universe continually procreates cluster generations \textit{of energy} that physically originate from earlier energy cluster generations. If the scaling of the universe's energy is uniform, then each succeeding cluster generation in the universe has the same amount of energy. By way of analogy, organisms procreate new generations \textit{of organisms} that physically originate from earlier organism generations. Procreation appears to follow the same model, for the universe, for organisms and for social networks. The physicist John Wheeler said  that 

\begin{quote}
	``the laws of physics \dots must have come into being at the big bang'' \cite{Wheeler1990}. 
\end{quote}

The physical environment of a social network precedes its existence. Suppose that an energy source distributes energy isotropically to a social network's physical environment. This supposition is a reasonable approximation to conditions at the surface of the Earth. For a small geographic area of a few square miles, for a height above ground of up to 100 feet say, and for a time period of a few hours or a day, the air temperature is usually contained within a moderate range. A uniform temperature would indicate that, approximately, heat energy is isotropically distributed within such an atmospheric volume. Similarly, energy received from the sun's radiation in such an area would be relatively uniform---isotropic. In order not to waste the available energy, the energy-receiving plants and animals should emulate an approximately isotropic hierarchical structure; otherwise there will be a mismatch between energy inputs and energy usage by the receivers, and available energy will be wasted. The isotropic distribution of energy appears to describe the ideally efficient manner of energy distribution in our universe, given the isotropic manner of its creation.

Suppose the universe's total energy is $\omega$, its initial energy density $E_{0}$, and its initial volume $V_{0}$ such that, for any $k^{th}$ cluster generation,

\begin{equation}\label{Isotropy eq 10.10}
\\ \frac{E_{0} }{S^{k}} \times [S^{k}V_{0}] = E_{0} \times V_{0}=\omega.
\end{equation}

The implication of (\ref{Isotropy eq 10.10}) for a universe to which it applies is that subsystems in the universe of the same size contain the same amount of energy distributed in a uniform way. This implication is analogous to \textit{The General Collective Problem Solving Capacity Hypothesis}: subsets of society's solved problems that contain the same amount of information required the same---uniform---amount of energy to create.

For the universe to which (\ref{Isotropy eq 10.10}) applies, the entropy $\eta$ of the universe is 

\begin{equation}\label{eq 43}
\\ \log_{S} \left( S^{k} \right)= k.
\end{equation}

In thermodynamics, entropy was originally defined with respect to its rate of \textit{change}, $d\eta$. The classical definition of entropy is

\begin{equation}\label{eq 44-8}
	 d \eta = \frac{dQ}{T}
\end{equation}
 
\noindent for $Q$ a quantity of heat, $dQ$ a change in that amount of heat, and $T$ the absolute temperature at which the change of heat occurs. Since $T \propto Q$, it could be said that the classical definition of entropy implies that the logarithm of energy density is inversely proportional to entropy, as an energy scaling perspective implies. In terms of the preceding discussion, the change in entropy is the logarithm of the ratio of the number of clusters for a later $k$th cluster generation compared to an earlier $j$th cluster generation, $k -  j > 0$, so that 

\begin{equation}\label{eq 44}
\\ d \eta = \log_{S} \left(\frac{ S^{k}}{ S^{j}} \right)= k-j.
\end{equation}

As $k$ increases, the number of clusters, $S^{k}$, per cluster generation increases, the volume of the universe increases, and the energy density of a cluster decreases. Here, $k - j > 0$ implies that the energy density of the universe is decreasing as the entropy increases, consistent with entropy being a measure of the increasing dispersion of energy in the universe. Entropy in this view is a way of counting the scalings---the number of $S$-scaled energy cluster generations---of the energy of the universe. If scaling occurs at a fixed rate, the rate of change in entropy is proportional to the rate of change in time, all other things being equal; entropy is proportional to the age of the universe. In this perspective, for the universe, 

\begin{equation}\label{eq 44-2}
	\eta \propto S^{k}
\end{equation}

\noindent for $k$ generations, and in general,

\begin{equation}\label{eq 44-4}
	 d \eta \propto \log_{S} \left(\frac{ S^{k}}{ S^{j}} \right)= k-j,
\end{equation}
 
 \noindent for $k -  j > 0$. In particular, using  (\ref{eq 44}), we may infer equality such that

\begin{equation}\label{eq 44-6}
	 d \eta = \log_{S} \left(\frac{ S^{k}}{ S^{j}} \right)= k-j .
\end{equation}

If the \textit{Capacity Multiplier Theorem} applied to the universe's energy, then the universe's fastest bi-nodal rate multiplied by the entropy of all the universe's energy would give a maximum speed limit for that universe, because no entropy of energy  could be greater the entropy of the universe's total energy. 

The energy scaling perspective on entropy is consistent with ideas about the big bang origin of the universe, ideas which did not exist until long after Boltzmann's probabilistic derivation of entropy in the 1870s. The energy scaling perspective  permits a succinct explanation of why the natural logarithm plays a prominent role in our universe. We can infer that the natural logarithm arises because the universe is, in a sense, an Ideal Network, that is, isotropic.

\subsubsection{The Natural Logarithm Theorem}

Suppose a single node is supplied with $S\epsilon$  energy units, which enables the node to \textit{transmit} information $S$ steps. With $S\epsilon$  energy units, the node can reach any node in the network. That is, following (\ref{eq 41}) and (\ref{eq 41-2}), in general the transmission \textit{capacity} of a single node using $S\epsilon$ energy units per time unit is 

\begin{equation}\label{eq 45}
	 S^{\eta}= n
\end{equation}

\noindent  nodes. The same transmitting node, using $S\epsilon$ energy units per time unit, can traverse $\eta = \log_{S} (n)$ cluster generations of networked nodes per time unit. 

The capacity \textit{of a node} in a cluster generation---each nested cluster generation contains all the network's nodes---to \textit{receive} information in one of the $\eta = \log_{S} (n)$ cluster generations is

\begin{equation}\label{eq 46}
\begin{split}
	 \frac{d \eta}{dn}&=\frac{ d \left[\log_{S}S^{\eta}\right]} {d(S^{\eta})}\\
&=\frac{1}{\ln\left(S\right){S^{\eta}}}.
\end{split}
\end{equation}\

In an isotropic network, the $S^{\eta} $ per node capacity to transmit information to the network's $n$ nodes should be exactly \textit{inverse} to the $\frac{1}{\ln\left(S\right){S^{\eta}}}$ per node capacity to receive information. That is, in an isotropic network 

\begin{equation}\label{eq 47}
\\ S^{\eta} =\ln\left(S\right){S^{\eta}}.
\end{equation}\

\noindent It follows from (\ref{eq 47}) that $\ln (S) = 1$. Thus, for an isotropic network, $S=e$; in an isotropic network, the path length is the natural logarithm, $e$. This result applies to isotropic systems generally.	That is, we have demonstrated

\begin{The Natural Logarithm Theorem}
An isotropic network is scaled by the natural logarithm $e$. That is, in an isotropic network the scaling factor (and the path length)
\begin{center}
$S=e$.
\end{center}
\end{The Natural Logarithm Theorem}

A corollary of \textit{The Natural Logarithm Theorem} is: if a network is isotropic, its path length is invariant.

\textit{The Natural Logarithm Theorem} provides an explanation of the role of the natural logarithm in our universe. If an isotropic network is the most energy efficient kind of network---the `Ideal Network'---then networks which obtain an advantage by being energy efficient should adapt to be approximately isotropic. As the cumulative amount of energy supplied to a network grows, the network should grow, according to \textit{The Natural Logarithm Theorem}, by a power of the natural logarithm. The natural logarithm's role in modeling many different kinds of systems, including the energy relationship between quanta in quantum mechanics,  is consistent with \textit{The Natural Logarithm Theorem}. 

The energy scaling perspective facilitates proof of \textit{The Natural Logarithm Theorem}. The identification of the path length as proportional to the energy scaling factor plays an important, if not indispensable, role in the proof of \textit{The Natural Logarithm Theorem}. The proof of \textit{The Natural Logarithm Theorem} using the energy scaling perspective increases our confidence that the path length plays a vital role in networks. It is consistent with the fundamental role that the natural logarithm plays in so many phenomena that a fundamental and intrinsic feature of networks, namely the path length, is required to show the fundamental and intrinsic role of the natural logarithm in isotropic networks. This proof is consistent with \textit{The Isotropic Network Rate Theorem} and the \textit{Capacity Multiplier Theorem}, and supports their validity.

\subsubsection{The Network Entropy Theorem}

Using The Natural Logarithm Theorem, we can now state 

\begin{The Network Entropy Theorem} The formula for the entropy $\eta$ of a network is in general
\begin{equation}
	\eta = C \log_S(n),
\end{equation}
where $C$ is the network's clustering coefficient, and $S$ is its path length. In particular, for an isotropic network, $C=1$, $S=e$ where $e$ is the natural logarithm, and 
\begin{equation}
	\eta = \log_e(n).
\end{equation}

\end{The Network Entropy Theorem}

\section{Testing a theory of intelligence}\label{Section, testing theory}

\begin{table}
\begin{center}
\small
	\begin{tabular}{|c|c|c|c|c|c|c|}\hline
	\textbf{Network} & Nodes  & Number of nodes & \textit{S} & \textit{C} & $\eta$ & Notes\\ \hline
Actors & people & 225,226 & 3.65 & 0.79 & 7.52 & {\footnotesize\ 1} \\ \hline
\textit{C. elegans} & neurons & 282& 2.65 & 0.28 & 1.62 &{\footnotesize\ 1}\\ \hline
Human Brain & neurons & $10^{11}$ & 2.49 & 0.53 & 14.71 & {\footnotesize\ 2}\\ \hline
1989 English & words & 616,500 & 2.67 & 0.437 & 5.932 & {\footnotesize\ 3, 4}\\ \hline
1150 English & words & 34,020 & 2.67 & 0.437 & 4.643 & {\footnotesize\ 4, 5}\\ \hline
1657 English & words  & 200,000 & 2.67 & 0.437 & 5.431 & {\footnotesize\ 4, 6}\\ \hline
1989 population & people & 350,000,000 &  3.65 & 0.79 & 12.0 & {\footnotesize\ 7, 8}\\ \hline
1150 population & people & 2,300,000 &  3.65 & 0.79 & 8.938 & {\footnotesize\ 7, 9}\\ \hline
1657 population & people & 5,281,347 &  3.65 & 0.79 & 9.445 & {\footnotesize\ 7, 10}\\ \hline
	\end{tabular}
	\caption{Calculations of $\eta$}\label{Table 2}
	\end{center}

\normalsize
\noindent \textbf{Notes to Table} \ref{Table 2} \\
\\
1.\ The network of actors, its number of nodes, and the values for $S$ and $C$ are from \cite{WS1998}.\\
2.\ The number of neurons is from \cite[p.480]{Nicholls2001}. $S$ and $C$ are from \cite{Achard2006}.\\
3.\ The number of words is from the OED. \\
4.\ The values for $S$ and $C$ are from \cite{Ferrer2001} based on about 3/4 of the million words appearing in the British National Corpus. The British National Corpus consists of about 70 million words of written English, used to obtain statistical information about the use of the English lexicon. \cite{Motter2002} found $S$=3.16 and $C$=.53 based on an online English thesaurus of about 30,000 words. \cite{Ferrer2001}'s sample size is larger and likely more representative of actual usage of English words.\\
5.\ The number of words is based on the University of Toronto's Old English Dictionary project.\\
6.\ The number of words is from EMEDD \cite{EMEDD}.\\
7.\ The values for $S$ and $C$ are based on the actors study of \cite{WS1998}.\\
8.\ The number of people is an estimate of the English speaking societies in 1989, determined by adding the number of people determined for censuses for the USA, 248.7 million people, according to the US census 1990 \cite{UScensus2000}; Canada 27,296,859 people in 1991 \cite{CanadaCensus}; England 50,748,000  people in 1991 \cite{UKcensus}; Australia, 16,850,540  people in 1991 \cite[p. 11]{AustraliaCensus}. These total 343,595,000 people.\\
9.\ The number of people in England is based on Hinde's remark (at p. 28) in his book on England's population that 1.6 to 1.7 million people at the time of the Domesday Book, 1086, are the estimates that are most likely to be accurate, and his estimate (at p. 24) of English population growth of 0.5\% per year \ for the period 1086 to 1348 \cite{Hinde2003}. On the Domesday population, similar estimates are found in \cite[p. 149]{Sawyer1998}, \cite[p. 53]{Grigg1980}, and \cite[p. 34]{Snooks1995}.\\
10.\ The number of people in England is from \cite[Table 7.8, following p. 207, for the year 1656]{Wrigley1989}. 
\end{table}

\subsection{About the data in Table \ref{Table 2}}

Thus far we have found, in addition to the principled arguments in favor of the existence of society's average general collective problem solving capacity, statistical support using the rate at which average IQs increase, the English lexicon grows and lighting efficiency improves. Having found an actual rate, $3.41\%$ per decade, for society's average general collective problem solving rate, (\ref{eq Networking-10.1000}) and (\ref{eq 24-Av}) suggest attempting to find the average individual's physiologically innate, or basal, problem solving rate. To find the average innate, or basal, problem solving rate led to the quest for functions of population and of society's solved problems that multiply the innate, or basal, problem solving rate \textit{in a network}, which, it is proposed, is $C \log_{S} (n)$. Principled arguments in favor of the validity of $C \log_{S} (n)$ as a network's rate multiplier have been set out above.  To increase our confidence that $C \log_{S} (n)$ multiplies the innate, or basal, problem solving rate \textit{in a network}, we require numerical results that are consistent with its proposed role as a network rate multiplier. The more unlikely it is that the numerical results are close to each other as a result of coincidence, the stronger our confidence that $f=g=C \log_{S} N(t)$. Each result that is consistent with \textit{The General Collective Problem Solving Capacity Hypothesis} may not by itself be proof of the existence of society's general collective problem solving capacity, but several different consistent results, obtained in different ways with different data, may together increase our confidence that \textit{The General Collective Problem Solving Capacity Hypothesis} is true. 
 
For the first through fifth sets of  numerical results obtained for the purpose of  supporting the role of $C \log_{S} (n)$ as a network's rate multiplier, I use the data set out in Table \ref{Table 2}. For each indicated category in Table \ref{Table 2}, the number of nodes has been determined from sources that are described in the notes to the table. In the column headed $S$ is the path length, in the column headed $C$, the clustering coefficient, for the different networks listed in the first column in Table \ref{Table 2}. 

The path lengths and clustering coefficients for English populations at the various times shown in Table \ref{Table 2} are based on the path length and clustering coefficient measured by Watts and Strogatz for 225,226 actors in the Internet Movie Database, set out in their 1998 article \cite{WS1998}. We assume, in the calculations that follow, that 225,226 actors are a large enough number of people to be representative of the English speaking networked population generally. Since we infer that social networks in similar societies all approximate a general standard---an Ideal Network---for social networks, and that culture changes slowly enough not to much alter the values for $S$ and $C$ for a society over time, the values for $S$ and $C$ in the study by Watts and Strogatz are likely similar to those for the different English populations described in the first column in Table \ref{Table 2}. 

Similarly, for the different English lexicons, we assume that the English lexicons conform to a standard criterion for lexical efficiency, and therefore we use, for the English lexicon for years 1150 and 1657, the $S$ and $C$, found, with the `improved method,' in the paper by Ramon Ferrer i Cancho and Richard V. Sol\'e \cite{Ferrer2001}. For each category in Table \ref{Table 2}, I calculated $\eta$  using  (\ref{eq 42}), based on the data in same row of Table \ref{Table 2}.

\subsection{The natural logarithm}\label{subsection nat log test}

The first set of data from Table \ref{Table 2} that we consider relates to \textit{The Natural Logarithm Theorem}. Based on \textit{The Natural Logarithm Theorem}, an isotropic information exchange network is predicted to have a path length close to the value of the natural logarithm \textit{e}, 2.71828, because, as set out in (\ref{eq 41}) and (\ref{eq 41-2}), the path length is strictly proportional to the energy scaling factor for an isotropic network. We would expect that a neural network, such as the human brain, which uses a large proportion of an organism's energy, and is vital to problem solving based on the organism's sensory input---information---should be structured to be approximately isotropic to maximize its energy efficiency, at least in relation to the network's path length. A similar observation applies to a network of solved problems, such as a lexicon. Therefore, we would expect that for a neural network and a network of solved problems affecting the survivability of an organism, the path length should be close to the path length of an isotropic network. The values in Table \ref{Table 2} for $S$ for the nervous system of the worm, \textit{C. elegans} (2.65), for the English lexicon (2.67), and for the human brain (2.49) are pretty close to the value of the natural logarithm, $e$, 2.71828, consistent with our principled expectations. Further, if \textit{The Natural Logarithm Theorem} is valid, we have gained some insight into the structure of these neural and abstraction networks; they likely have information exchanges that are approximately isotropic, consistent with our expectations. 

In a \textit{social network}, not all personal---bi-nodal---relationships are equal; a social network of human beings is not isotropic. The  contrapositive of \textit{The Natural Logarithm Theorem} is that if the value  of $S$ is not equal to the natural logarithm, then the network is not isotropic. In Table \ref{Table 2}, the value of $S$ for the network of 225,226 actors studied in \cite{WS1998}, 3.65, is not as close to the value of the natural logarithm as for a neural network, consistent with our expectations. \textit{The Natural Logarithm Theorem}, which arises out of \textit{The General Collective Problem Solving Capacity Hypothesis} and \textit{The Isotropic Network Rate Theorem}, appears to predict the path length for isotropic networks. We have consistency with \textit{The General Collective Problem Solving Capacity Hypothesis}. The values of the path lengths for neural networks and a network of abstractions, if we assume that such systems are isotropic, are consistent with the path lengths predicted by \textit{The Natural Logarithm Theorem}. 

\subsection{The human brain}\label{subsubsec Brain}  

The second set of data relates to the problem solving capacity of the human brain. In vertebrates, larger nerve fibers are myelinated \cite[p. 123]{Nicholls2001}. The conduction velocity of myelinated nerve fibers varies from ``a few meters per second to more than 100 m/s'' \cite[p. 123]{Nicholls2001}. The measured conduction velocity is a pair-wise, or bi-nodal, rate of transmission. To calculate the \textit{networked} conduction velocity for the average nerve fiber in the human brain, we apply the \textit{Capacity Multiplier Theorem} and multiply the average conduction velocity by the $\eta $ of the human brain's networked neurons  estimated in Table \ref{Table 2} to be 14.71. In other words, the brain's network transmission rate can be estimated to be, on average, 14.71 times as fast as the average pair-wise, or bi-nodal, conduction velocity of one of its nerve fibers. Such a network transmission rate should confer on the human brain a problem solving capacity and speed that exceeds what we might expect by just looking at the bi-nodal velocity of a nerve fibre. If the \textit{Capacity Mulitiplier Theorem} is correct, it may help explain the observed capacities of the human brain. This expectation is consistent with what we know about human problem solving capacities, but it is not proof of \textit{The General Collective Problem Solving Capacity Hypothesis}.

\subsection{The innate, or basal, rate of problem solving}\label{Subsubsec Basal rate}

The third set of data has to do with calculation of the average individual's innate, or basal, problem solving capacity. In the earlier part of this article, we estimated that society's average general collective problem solving rate is  3.41\% per decade. Relying on \textit{The General Collective Problem Solving Capacity Hypothesis}, the lexicon is a set of solved problems that increases in size at the same rate as the set of all of society's solved problems. We will use the English lexicon and population data to estimate the value of $\eta$ for the number of solved problems and population at different years, using the data in Table \ref{Table 2}.  We found in  (\ref{eq 36}) that

\begin{equation}\label{eq 48}
	 \frac{d |\{PS_{Society}\}|}{dt} 
= \left(\frac{d|\{ PS_{Innate}\}|}{dt}\right)_{Av} \times \eta(Pop) \times \eta(Lex),
\end{equation}

\noindent where 

\begin{equation}\label{eq 48-2}
	 \frac{d |\{PS_{Society}\}|}{dt} = 3.41\%
\end{equation}

\noindent  per decade is society's average general collective problem solving capacity. We assume that for the past few thousand years, the average innate, or basal rate, of problem solving, 

\begin{equation}\label{eq 48-4}
	\left(\frac{d |\{PS_{Innate}(t)\}|} {dt}\right)_{Av}=m,
\end{equation}

\noindent the first factor on the right side of (\ref{eq 48}), has been constant. We need to be able to assign numerical values to $\eta(Pop)$ and $\eta(Lex)$, the other two factors on the right side of (\ref{eq 48}).

We have calculated  $\eta(Pop)(t)$ in Table \ref{Table 2} for the English speaking population at the years 1150, 1657 and 1989 and $\eta(Lex)(t)$ for the English lexicon at the years 1150, 1657 and 1989. Using this information, we are now in a position to calculate an estimate of

\begin{equation}\label{eq 48-6}
	\left(\frac{d |\{PS_{Innate}(t)\}|}{dt}\right)_{Av}.
\end{equation}

\noindent Since

\begin{equation}\label{eq 48-8}
	 \frac{d |\{PS_{Society}\}|}{dt} = 3.41\%
\end{equation}

\noindent per decade is an \textit{average rate} over time, the two factors $\eta ( Pop( t ) )$ and $\eta( |Lex (t)|) $ must also be averages  for the period over which the rate is calculated. To calculate the product of $\eta ( Pop) \times \eta ( Lex )$ for a time period $t_{1}$ to $t_{2}$, we need to calculate, for network $X$,

\begin{equation}\label{Eq Averaging-eta20.10}
	\frac{ \eta(X(t_{1})) + \eta(X(t_{2}))}{2} , 
\end{equation}

\noindent for each of the population and the lexical networks, during the relevant period to get the average applicable values of $\eta$. We take the average of the logarithmic functions since $C \log_{S} (n)$ changes proportionally with time. Using the results about $\eta$ from Table \ref{Table 2}, we obtain average values as set out in Table \ref{Table 3}.

\renewcommand{\arraystretch}{1.25}
\begin{table}[ht]
\begin{center}
\footnotesize
	\begin{tabular}{|c|c|c|c|c|}\hline
	$t_{1}$ & $t_{2}$ & Average $\eta ( Pop)$ & Average $\eta ( Lex )$&Product of 2 averaged $\eta$s \\ \hline
1150 & 1989 & 10.47 & 5.29 &55.37\\ \hline
1657 & 1989 & 10.72& 5.68 &60.94\\ \hline
	\end{tabular}
		\caption{Calculation of average values of $\eta$s and their products, for 1150--1989 and 1657--1989.} \label{Table 3}
		\end{center}
\end{table}

\normalsize
Using the data in Table \ref{Table 3}, we can find 

\begin{equation}\label{eq 48-10}
	\left( \frac{d |\{PS_{Innate}(t)\}|}{dt} \right)_{Av} ,
\end{equation}

\noindent by solving for it using (\ref{eq 48}), as follows: 

\begin{equation}\label{eq 48-12}
\begin{split}
	 \left( \frac{d |\{PS_{Innate}(t)\}|}{dt} \right)_{Av}
&=  \left(\frac{d |\{PS_{Society}(t)\}|}{dt}\right) _{Av} \\
& \div \left\{ \eta([Pop(t)]_{Av}) \times \eta( |Lex(t) |_{Av})\right\}\\
&=3.41\% / \ decade\\
& \div \left\{ \eta([Pop(t)]_{Av}) \times \eta( |Lex(t) |_{Av})\right\}.
\end{split}
\end{equation}

\noindent We use the value in subsection \ref{Subsection inf and problems} for the average collective problem solving rate $\frac{d |\{PS_{Society}\}|}{dt}$, 3.41\% per decade, and the product of the averaged values for $\eta(N(t))$ in Table \ref{Table 3} for the periods 1150--1989 and 1657--1989, to calculate the average innate problem solving rate set out in Table \ref{Table 4}. In using the product of the averaged values for $\eta(N(t))$ in Table \ref{Table 3}, the size of the English lexicon at 1150 may be understated, and thus may result in a slight under-calculation of $\eta(N(1150))$. If $\eta(N(1150))$ calculated for the English lexicon is slightly lower than its actual value, that would result in a too high general collective problem solving rate for the period 1150--1989. The estimate based on the average $\eta$ values for population and lexicon for the period 1657--1989, and using the rate of increase in society's problem solving rate as 3.41\% per decade, is 5.60\% per thousand years. 

\renewcommand{\arraystretch}{2.5}
\begin{table}[ht]
\begin{center}
	\begin{tabular}{|c|c|c|}\hline
	$t_{1}$ & $t_{2}$ & $\frac{d[ PS_{Average \ Innate}(t)]}{dt}$\\ \hline
1150 & 1989 & 6.16\% per 1000 yrs\\ \hline
1657 & 1989 & 5.60\% per 1000 yrs\\ \hline
	\end{tabular}
		\caption{Calculation of the average innate problem solving rate, based on an average general collective problem solving rate of 3.41\% per decade}  \label{Table 4}
		\end{center}
\end{table}

\subsubsection{Comparing values of $\eta$}\label{comparing eta}

The product of  $\eta( Pop(1989) )$ and $\eta( |Lex (1989)|)$ for the English speaking population and for the English lexicon, using the values in Table \ref{Table 2} is $12 \times 5.932 = 71.21$. The average innate, or basal, problem solving capacity of an individual in English speaking society was multiplied 71.21 times compared to what the average innate problem solving rate would be without any social networking at all, not even with parents, and without any language at all, not even animal-like sounds or gestures. Instead of comparing the average problem solving rate of a modern individual to a person bereft of parents and language, let's use as a basis for comparison a physiologically modern person living in a primitive society of 150 individuals with 100 vocalizations. If we use modern values of $C$ and $S$ for $\eta (Pop)$ for the primitive population and for $\eta(| Lex| )$ for their lexicon, we find $\eta(Pop) = 3.057$ and $\eta( |Lex| )= 2.049$; the $\eta$ product of $\eta(Pop=150) \times \eta( |Lex = 100|)$ is 6.265. The product of $\eta(Pop(1989))$ and $\eta( |Lex(1989)|)$ is \textit{11.37 times larger} than the $\eta$ product, 6.265, of the 150 member society with 100 vocalizations. We can consider the $\eta$ product for the primitive society, 6.265, to approximately represent the capacity multiplier for individuals in a `primitive' society of physiologically modern humans. If the lexicon had 10,000 words, then $\eta( |Lex| )= 4.098$ and the $\eta$ product for the same size society of 150 individuals would be 12.53, twice as much. Language multiplies the problem solving capacity of human beings. 

To get a sense of the difference that large-scale social networking and a large network of solved problems makes in developing individual intelligence, consider that the $\eta$ -value of the human brain, as set out in Table \ref{Table 2}, 14.71, is 9.08 times that of the $\eta$-value in Table \ref{Table 2} of the neural system of the worm C elegans, which is less than the 11.37 times difference between a modern society's capacity multiplier for a physiologically modern person compared to a a primitive society's capacity multiplier for a physiologically modern person living in a primitive society.  There was more difference between the problem solving capacity multiplier of English speaking society in 1989, considered as a single mind, and the problem solving capacity multiplier of a primitive society of 150 people with only 100 words, than there is between the brain of an average human being and the nervous system of the worm, C. elegans.

This third set of data is consistent with what we know about human nature and human capacities, but it is not proof of \textit{The General Collective Problem Solving Capacity Hypothesis}. 

\subsection{Glottochronology}

The fourth set of data also relates to the average innate, or basal, problem solving capacity of an individual. We have just found estimates, set out in Table \ref{Table 4}, of the average innate problem solving rate. Using a second, entirely different method, using entirely different data, we estimate the average innate, or basal, problem solving rate of a modern human being. We then compare the two differently obtained estimates of the average innate, or basal, problem solving capacity of a human being. 

In the 1940s, the linguists Morris Swadesh and Robert Lees devised a way to estimate how long ago two related languages first diverged. The method is called glottochronology. The method used in glottochronology was this. Swadesh chose 100 words (the `Basic List') that he thought were least likely to change in a language; it is not possible to study the historical usage of words over a long period of time if the words only appeared in the language for a few years. Swadesh studied actual recorded historical uses of cognates (related words) on the Basic List,  for two related languages. Swadesh measured the rate of retention of cognates in related languages by counting the current number of Basic List words they still shared: ``a maximum \textit{retention} of 90 percent after a thousand years, a minimum of 81 percent, and an average of 86 percent'' \cite[p. 276]{Swadesh1971}. If we convert the rate of retention to a rate of divergence, then Swadesh's estimated rate of divergence ranged from 10\% to 19\% per thousand years, an average of 14\% per thousand years. Using the rate of divergence of the words on the Basic List, one can estimate the date of the common ancestral language. 

Using his idea, glottochronology, Swadesh estimated that English's ancestral language began ``at least seven thousand years before the present,'' \cite[p. 84]{Swadesh1971}, probably using the average rate of divergence that he calculated. His book was published posthumously; the `present' was about 1966. Swadesh did his work before the advent of the personal computer. In 2003, 37 years later after his 1966 estimate of the age of English's ancestral language of ``at least seven thousand years before the present,'' Russell D. Gray and Quentin D. Atkinson estimated that Indo-European (English's ancestral language) began 8700 years ago,  using newer, more sophisticated methods \cite{Gray2003}. To adjust Swadesh's estimated 14\% rate of divergence using Gray and Atkinson's more recent and comprehensive estimate, we need to adjust Swadesh's estimated average rate of language divergence, 14\% per thousand years, so that the adjusted rate of divergence dates Indo-European to 8700 years before Gray and Atkinson's article. To do so, we multiply 14\% per thousand years by 7037/8700, for an \textit{Adjusted Swadesh Rate} of divergence of 11.32\% per thousand years, as in

\begin{equation}\label{eq 49}
	\frac{7037}{8700} \times 14\% = 11.32\%.
\end{equation}

We now examine the relationship of the Adjusted Swadesh Rate of divergence of two related languages to the average innate problem solving rate. Suppose that two daughter languages share a common ancestral language with a lexicon $Lex_{Anc}$. Let $Lex_{D1}$ represent the lexicon of the first daughter language and $Lex_{D2}$ represent the lexicon of the second daughter language. Suppose the common ancestral language $Lex_{Anc}$ has a general collective problem solving rate $R$. At the time immediately before the two daughter languages begin to diverge from their common ancestral language, the general collective problem solving rate for each is \textit{the same}, namely $R$, and their lexicons, $Lex_{D1}$ and $Lex_{D2}$, are identical to $Lex_{Anc}$. After the societies speaking the daughter languages have separated, each daughter society \textit{independently}  applies the same general collective problem solving rate, $R$, to devise lexical solutions---different lexical solutions due to the physical, cultural and other divergences of the daughter societies---to its language problems. To keep the analysis uncomplicated, we suppose that the number of words in the daughter languages remains the same as in the ancestral language, unchanged throughout the period of divergence. Each daughter language diverges---grows away---from the ancestral language \textit{at the same rate}, $m$, say, a percentage rate per thousand years. If the two daughter societies find different solutions for language problems, then on average they diverge from each other at the rate 2$m$, like the equal sides of an isosceles triangle diverging from the common vertex. 

\emph{Half the modified 11.32\% per thousand years of the Adjusted Swadesh Rate, is 5.66\% per thousand years.}

The 5.66\% per thousand years using Swadesh's glottochronology is remarkably close to the value calculated for the innate, or basal, problem solving rate in Table \ref{Table 4}, 5.60\% per thousand years, using data for an English speaking society for the period 1657 to 1989, the formula (\ref{eq Networking-10.1000}) and the average general collective problem solving rate of 3.41\% per decade. Since the data for 1657 is likely somewhat more reliable than our data for 1150 (the estimated English lexical growth rate for 1657--1989 is closer to the benchmark rate of increase in lighting efficiency than the estimated average English lexical growth rate for 1159--1989), the estimate of the average innate, or basal, problem solving rate for the period 1657--1989 is somewhat more reliable than our estimate of that rate for the period 1150--1989. It seems that the estimates of the average innate, or basal, problem solving rate in Table \ref{Table 4} is measuring half the rate of divergence of two related languages. 

Both methods---using glottochronology and using the formula (\ref{eq Networking-10.1000})---so different from each other, give us estimates that are about the same for the average innate, or basal, problem solving rate of modern human beings. The numerical correspondence implies that the two different methods have measured the same phenomenon. 

This numerical correspondence, however, was obtained in a way that seems to contain an inconsistency. We supposed that the common ancestral language, and therefore each of the two daughter languages, had the same  general collective problem solving rate $R$. We inferred that the daughter languages diverge at twice their individual rate of divergence from the now fossilized ancestral language. But if each daughter language changes at the rate $R$, shouldn't $2m$ = $2R$? If so, then the rate of divergence would be twice 3.41\% per decade, that is 6.82\% \textit{per decade}, instead of the calculated 5.66\% (or so) per thousand years. Let's call this the `\textit{Divergence Rate Problem}.'

To resolve this apparent inconsistency, we require the \textit{Capacity Multiplier Theorem} in subsection \ref{Subsection, Net Rate Thm}. In the discussion about that theorem we observed that the average innate, or basal, problem solving  rate is unchanging. The reason why the Network Rate is a multiple of the innate, or basal, problem solving  rate is because the innate, or basal, problem solving rate is applied to $\eta$ cluster generations of solved problems.

Let $\eta_{Anc}$ here denote the product of $\eta(Pop)$ and $\eta (|Lex|)$ for the ancestral language. Let $m$ denote  the average innate, or basal, problem solving  rate. The number of word problems the society in which the ancestral language was spoken is equal to the average problem solving rate $m \times \eta$ which is applied to the existing number of solved lexical problems. That is, $m$ is applied so that we have $ m \times \eta \times |Lex|$. The \textit{Capacity Multiplier Theorem} stipulates that really it is the innate, or basal, problem solving rate that is applied to the networked set of society's lexical problems. The target of the ancestral society's  and a daughter society's problem solving capacity is the same scaled  number of problems, $\eta \times |Lex|$.

When the two daughter languages begin to diverge from each other, the ancestral language, now no longer spoken, has ceased changing. For the unchanging ancestral language, $m_{Anc}=0$. But for the daughter languages, the value of $m_{D1}= m_{D2}=m$. To find the number of changes in a  daughter language such as  $Lex_{D1}$ compared to the now \textit{unchanging}  ancestral language, we adapt the equation relating to solved problems set out in (\ref{eq A-80.10}) to the situation of a changing lexicon as follows:

 \begin{equation} \label{eq Lex-10.10}
	 |Lex_{D1}(t_1)| = \left\{(1+ m)^{ \Delta t}\right\} \times \eta \times |Lex_{D1}(t_0)|.
\end{equation}

Now we examine one period of time, so $\Delta t = 1$ at the beginning of divergence, when $Lex_{D1}(t_0) =
Lex_{Anc}(t_0)$. Since we are comparing the unchanging ancestor language at $(t_0)$ to the changing daughter language, $m_{Anc}=0$ for this purpose. We then take the ratio 

\begin{equation}\label{eq Lex-10.20}
\begin{split}
	 \frac{|Lex_{D1}(t_1)|}{|Lex_{Anc}(t_1)|}
&= \frac{\left\{(1+ m_{D1})^{ \Delta t}\right\} \times \eta \times |Lex_{D1}(t_0)|}{\left\{(1+ m_{Anc})^{ \Delta t}\right\} \times \eta \times |Lex_{Anc}(t_0)|}  \\
&= \frac{\left\{(1+ m_{D1})^{1}\right\} \times |Lex_{D1}(t_0)|}{\left\{(1+ 0)^{1}\right\} \times |Lex_{Anc}(t_0)|}  \\
& =1+ m_{D1} \\
& =1+ m 
\end{split}
\end{equation}\\

\noindent because $|Lex_{D1}(t_0)| =|Lex_{Anc}(t_0)|$ and $\eta_{D1}(t_0) =\eta_{Anc}(t_0)$. The last line in (\ref{eq Lex-10.20}) gives us the average innate problem solving rate $m$ in $1+ m$. Glottochronology indirectly measures the innate, or basal, problem solving rate---the average rate of divergence of two daughter languages is twice the average innate problem solving rate---because the innate problem solving rate is applied to $\eta$ times the number of solved lexical problems both for a daughter language and its ancestral language. 

Since the innate problem solving rate is almost the same when calculated using the formula for the entropy of a network and when determined using glottochronology, we conclude
\begin{enumerate}
	\item The validity of \textit{The General Network Rate Theorem} is consistent with this concurrence. 
	\item \textit{The General Network Rate Theorem} can be applied to perform actual calculations for networks. 
	\item The  \textit{Capacity Multiplier Theorem} is required to explain the concurrence of half the average rate of divergence for two related language as determined using glottochronology with the average innate, or basal, problem solving rate as determined using the concept of a network's entropy.
	\item Despite various shortcomings that have been identified in glottochronology's methodology \cite[p. 311]{Blust2000} and \cite[p. 204 and 205]{Campbell1998}, in principle, glottochronology indirectly measures the average  innate, or basal, problem solving rate (by measuring twice the innate, or basal, problem solving rate). The underlying concepts and methods of glottochronology are, in principle, valid.
\end{enumerate}

When, in 2007, I first attempted to resolve the \textit{Divergence Rate Problem}, I assumed that $ \eta \times m$ applied to the lexicon. That, in addition to some erroneous data, calculations and assumptions, led me to erroneously infer that there was a 4:1 ratio instead of the actual 2:1 ratio between the rate of divergence for daughter languages as compared to the innate rate.  

To the observation of such a close correspondence between 5.66\% per thousand years, which is half the 11.32\% per thousand years of the Adjusted Swadesh Rate of divergence of two  related languages, and the 5.6\% per thousand years calculated using $\eta$ and the data above, one can adopt an 1850 remark of R. Clausius about a result in thermodynamics:
\begin{quote}
	``Such an agreement between results which are obtained from entirely different principles cannot be accidental; it rather serves as a powerful confirmation of the two principles and the first subsidiary hypothesis annexed to them \cite{Clausius1850}.''
\end{quote}

The fourth set of data is consistent, and (I would submit) confirmatory. If so, Swadesh's glottochronology should be recognized for its role not only in linguistics but as an independent means of indirectly measuring and verifying society's average general collective problem solving rate.

\subsection{When language began}\label{subsubsec Test, lang begins}

The results concerning glottochronology can be generalized. Using the average innate, or basal, rate of growth for a system (not just a lexicon), it is possible to estimate the age of the system. The average innate, or basal, rate, $m$, can be used to find the length of time between the current system and its commencement, if we assume that at its commencement, $\eta=1$. We can do this by finding how long it would take, going back in time, for a system with $N(t_{2})$ nodes to reach an ancestor system of $N(t_{1})=S$ nodes, at the rate $m$. We assume that when $N(t_{1})=S$, then $\eta = 1$. This `entropy' dating method should be valid if the rate at which the entropy $\eta$ of the system changes is proportional to time. It must be appropriate to assume that the energy supply has been \textit{constant} during the time when the daughter system diverged from the parent system; if the system's energy supply---energy input---has been constant, then we may infer that the system's output should be constant as well, proportional to the energy input. We use $m$ as the rate in the equation (\ref{eq 9}). Instead of solving for the rate $r$ in (\ref{eq 9}), as we did in (\ref{eq 10}), we solve for $t_{1}$ in (\ref{eq 9}). 

Changing (\ref{eq 9}) slightly, we have ,

\begin{equation}\label{eq 51}
	 N(t_{2})=N(t_{1})\exp(m\cdot(t_{2}-t_{1})),
\end{equation}

\noindent where $t_{2}$ represents the current year, $N(t_{2})$ represents the number of nodes in the system at $t_{2}$, $m$ represents the basal rate of change, and the starting year $t_{1}$ is unknown. In (\ref{eq 51}), ${N(t_{1})}$ represents the size of the system at the year $t_{1}$. Solving (\ref{eq 51}) for $t_{1}$ gives

\begin{equation}\label{eq 52}
	 t_{1}=t_{2} -\left[\left\{\log \left(\frac{N(t_{2})}{N(t_{1})}\right)\right\} \div m\right].
\end{equation}

\noindent In (\ref{eq 52}), the expression

\begin{equation}\label{eq 53}
  \left[\left\{\log \left(\frac{N(t_{2})}{N(t_{1})}\right)\right\} \div m\right]
\end{equation}

\noindent gives the number of years before $t_{2}$.

This can be described as follows.

\begin{The Basal Rate Network Dating Theorem}
Suppose the innate, or basal rate, of growth in a network, $m$, is known. If $t_{2}$ represents a later year, $N(t_{2})$ represents the number of nodes in the network at $t_{2}$, and the unknown sought to be determined is an earlier year $t_{1}$, when it is estimated the network had ${N(t_{1})}$ nodes, then if $m$ is assumed to be constant for the period from $t_{1}$ to $t_{2}$,

\begin{equation}\label{eq 54}
	t_{1}=t_{2} -\left[\left\{\log \left(\frac{N(t_{2})}{N(t_{1})}\right)\right\} \div m\right] .
\end{equation}

\end{The Basal Rate Network Dating Theorem}

Using \textit{The Basal Rate Network Dating Theorem}, and the estimated average innate, or basal, problem solving rate, we can attempt to create a fifth set of data that estimates when language began. The average innate, or basal, problem solving rate, 5.66\% per thousand years, based on the Adjusted Swadesh estimate of the divergence rate for two related languages, is preferred because it is so close to the 5.60\% rate set out in Table \ref{Table 4}, determined using \textit{the General Network Rate Theorem}. For comparison purposes, I have also set out calculations using the less likely 6.16\% per thousand year rate  in Table \ref{Table 4}. Two  assumptions about the number of words at the beginning of language are possible. In one approach, we could assume that there was one word at the beginning of language, ${N(t_{1})} =1$. But primates and other animals have vocalizations which, while they may not be words as we think of them, communicate information. It may be more conservative to assume that at the outset of language, the precursors of modern human beings had some vocalizations, which I here assume to be 100, that is, ${N(t_{1})} =100$. I have shown both assumptions in Table \ref{Table 5}. Table \ref{Table 5} sets out estimates of when language began showing the two different assumptions, with one column headed by ${N(t_{1})} =1$ and the other by ${N(t_{1})} =100$, and using, at the left, two different values, obtained for the average innate problem solving rate, which are in Table \ref{Table 4}.

\begin{table}[ht]
\begin{center}
	\begin{tabular}{|c|c|c|c|}\hline
	 \multicolumn{2}{|c|}{Innate rate per 1000 years }&
	 \multicolumn{2}{c|}{Years before 1989}\\\cline{1-4}
	Source&Rate& ${N(t_{1})}=1$ & ${N(t_{1})}=100$\\ \hline
Adjusted Swadesh&5.66\% & 235,544 & 154,181\\ \hline
English 1150--1989&6.16\% & 216,425 & 141,666\\ \hline
	\end{tabular}
	\caption{ Estimating when language began based on the innate problem solving rate}  \label{Table 5}
		\end{center}
\end{table}

\normalsize

Estimating the start date of language is difficult because writing probably only began about 3500 B.C.E. \cite{Schmandt1992}. Speech leaves no records. A striking observation based on the data in Table \ref{Table 5} is the difference in the estimated starting date for language, depending on which of the two assumptions applies, ${N(t_{1})} =1$ or ${N(t_{1})} =100$. This may be taken to be indicative of how long it would take to build from one word to a 100 word lexicon at the estimated average general innate problem solving rate of about 5.66\% per thousand years---over 70,000 years---an indication of the amount of energy and time required to identify and solve new problems, or to discover new ideas. If the calculations---as opposed to the estimates---in Table \ref{Table 5} are accurate, they likely underestimate the age of language, because we have assumed that the average innate, or basal, problem solving rate was at all times constant. It is not unlikely that 150,000 years ago or more, the average innate problem solving rate was somewhat less than the current average innate problem solving rate because the human brain had a smaller physiological capacity, which would suggest the development of language would have taken somewhat longer than the estimates calculated in Table \ref{Table 5}. 

Mark Pagel did a study in which he calculated that seven slowly evolving words have half lives estimated at 166,000 years: I, we, who, two, three, four, five \cite[p. 205]{Pagel2000}. He remarks that ``These figures \ldots should not be taken literally'', but his results are not inconsistent with the estimates in Table \ref{Table 5}.

The estimates in Table \ref{Table 5}, a speculative exercise, are plausible, and perhaps indicative of when language began,  but there is no current data on language to corroborate them, though paleontological evidence is not inconsistent with such estimates. For example, human cooperative hunting seems to have begun about 500,000 years ago \cite[pp. 393--401]{Deacon1997}. This data gives no proof of \textit{The General Collective Problem Solving Capacity Hypothesis}. 

If \textit{The Basal Rate Network Dating Theorem} can be applied to date a process, such as evolution of an organism or of a structure in an organism, or of an economy, that permits the date to be corroborated, that would help support the validity of the ideas in this article. For example, can the time when the first neurons began be estimated---taking the number of neurons in a human brain at time $t_2$ and comparing that to one third the number of neurons in an ancestor when the brain was one third the size---using `entropy dating' and comparing the results, calculated for an earlier time, to paleontological evidence? 

\subsection{U.S. economic growth rate}\label{Subsubsec Test Ec growth}

The sixth set of data concern an economic calculation.  Earlier, in (\ref{eq 25}), we estimated that society's average collective problem solving rate---the rate of increase in society's store of solved problems---is 3.41\% per decade. We also found that the average individual problem solving rate is the same as society's average collective problem solving rate. The average U.S. \footnote{An `open' society such as the United States may be considered to provide simpler data for these purposes.} (for example) economic growth rate from 1880 to 1980---the rate of increase in society's store of material wealth---is much higher than 3.41\% per decade. Is this \textit{Economic Growth Rate Problem} inconsistent with the results this far?

Assume that economic productivity per person benefits from \textit{economic} networking. For a society with language and technology, suppose that the \textit{economic productivity} of an individual in a society results from the individual's problem solving capacity and networking with all people participating in the economic productivity of that society. Let 

\begin{equation}\label{eq 55}
	 \frac{d(Ec_{Pr}(t))}{dt}
\end{equation}

\noindent represent the  growth rate of productivity in an economy, $\eta (Pop)$ the entropy of the economy's population, and $(LP)$ the economy's labor participation rate. We infer the following:

\begin{equation}\label{eq 56}
	 \frac{d(Ec_{Pr}(t))}{dt}=\left( \frac{d |\{PS_{Individual} \}|}{dt}\right)_{Av} \times \left[\eta (Pop)\right]_{Av} \times(LP)_{Av} .
\end{equation}

We can use (\ref{eq 56}) to estimate the rate of economic growth for the United States from 1880 to 1980. To do so, we will use the following additional data:

\begin{itemize}
	\item •In 1880, the U.S. Census Office counted 50,155,783 people \cite[Table Ia]{1880UScensus}. $\eta(50,155,783) = 10.818657$, using the values for $C$ and $S$ found for the 225,226 IMDB actors in \cite{WS1998}.
	\item •In 1980, the U.S. Census Bureau counted 226,545,805 people \cite[Table 72]{1980UScensus}. $\eta(226,545,805) = 11.738675$, again using the values for $C$ and $S$ found for the 225,226 IMDB actors in \cite{WS1998}.
•	\item For 2004--2005, the U.S. labor participation rate was an estimated 66\% \cite{Mosisa2006}. We will assume this percentage applied in both 1880 and 1980. (It was estimated to have peaked at 67.1\% in the late 1990s \cite{Mosisa2006}.)
•	\item The average of the two $\eta$ values for the U.S. population, for 1880 and 1980, is 11.485141.
\end{itemize}

Using the average $\eta$ for population, and calculating using (\ref{eq 56}), 

\begin{equation}\label{eq 57}
\begin{split}
 \frac{d(Ec_{Pr}(t))}{dt}&=\left( \frac{d |\{PS_{Individual}\}| }{dt}\right)_{Av} \times \left[\eta (Pop)\right]_{Av} \times(LP)_{Av}\\
 &=3.41\% \ per \ decade \times 11.485141 \times 0.66\\
&=2.53\% \ per \ year 
\end{split}
\end{equation}

In fact, productivity per hour increased in the United States about 10 times over a 100 years up to about 1980, an average rate of about 2.3\% per year \cite{Romer1990}. 

The networked productivity growth rate, 2.53\% per year, calculated using (\ref{eq 55}) is 10.36\% higher than the 2.3\% per year based on \cite{Romer1990}. The discrepancy could result from the 10 times increase being only approximate, from inaccurate data or methodology, or from a missing factor represented by the function $h$---the infrastructure function---described in connection with (\ref{eq 28}), or could reflect some inefficiency in the market. The discrepancy could result from the informal economy. The informal labor force in Los Angeles County in 2004, based on the midpoint of a range of estimates, is 15\% \cite{Vogel2006}. A world bank study estimated the informal economy in 2000 in the U.S. at 8.8\% and in Canada 16.4\% \cite{Schneider2002}, so it is possible that the discrepancy of 10.36\% indirectly measures the informal economy. The short duration used to estimate the average rate of productivity growth may result in some uncertainty in the estimate; measurement over a 100 year period has more uncertainty than measurement over a longer period, as shown in Tables \ref{Table 1-1} and \ref{Table 1-1Lower}. In any event, (\ref{eq 56}) seems to address, at least partly, the \textit{Economic Growth Rate Problem}. This result is consistent with the results thus far, but not proof of \textit{The General Collective Problem Solving Capacity Hypothesis}.

\subsubsection{Economic Productivity Theorem}

From (\ref{eq 56})

\begin{equation}\label{eq 58}
	 \frac{d(Ec_{Pr}(t))}{dt}\propto  \left(\frac{d |\{PS_{Individual}\}|}{dt}\right)_{Av} \times \left[\eta (Pop)\right]_{Av}.
\end{equation}

\noindent Based on (\ref{eq 24-Av}),

\begin{equation}\label{eq Ec.24-Av}
\begin{split}
	 & \left(\frac{d |\{PS_{Individual}\}|}{dt}\right)_{Av}\\
 & \propto [C(PS_{Soc} )]_{Av} \times \log_{[S-PS](Av)}(|\{PS_{Society}\}|) \\
 & \times \left[C(Pop)\right]_{Av} \times \log_{[S-Pop](Av)}(Pop) ,
\end{split}
\end{equation}

\noindent or, more succinctly,

\begin{equation}\label{eq Ec.30-Av}
	  \left(\frac{d |\{PS_{Individual}\}|}{dt}\right)_{Av} \propto \left[\eta (|\{PS_{Society}\}|)\right]_{Av} \times \left[\eta (Pop)\right]_{Av} .
\end{equation}

\noindent If we substitute the right side of (\ref{eq Ec.30-Av}) into (\ref{eq 58}) for the factor 

\begin{equation}\label{eq Ec.40-Av}
\left(\frac{d |\{PS_{Individual}\}|}{dt}\right)_{Av},
\end{equation}

\noindent we obtain

\begin{equation}\label{eq Ec.50-Av}
	 \frac{d(Ec_{Pr}(t))}{dt}\propto  \left( \eta (|\{PS_{Society}\}|)_{Av} \times \eta (Pop) \right)_{Av} \times \left[\eta (Pop)\right]_{Av},
\end{equation}

\noindent and so, in general terms (leaving out the `Av' subscripts),

\begin{equation}\label{eq 61}
	 \frac{d(Ec_{Pr})}{dt}\propto \left\{\eta (Pop)\right\}^{2} \times \eta(|\left\{PS_{Society}\right\}|).
\end{equation}

\noindent This is an 

\begin{Economic Productivity Theorem}
Economic productivity is proportional to the square of the entropy of a society's population times the entropy of the number of the society's solved problems. 
\end{Economic Productivity Theorem}

\textit{The Economic Productivity Theorem} implies that impairing the capacity of a society's members to freely network and exchange ideas can reduce the value of the product on the right side of (\ref{eq 61}). The benefit of networking is increased---$\eta(Pop)$ is increased---when all of society's members can freely exchange solved problems. Since $\eta(Pop)$ by itself can affect the capacity of individuals and of a society, and therefor economic productivity, the value of $[\eta(Pop)]^{2}$ in relation to a society's economic growth is greater than that. If the freedom of a population to socially and economically network is impaired, then likely, for $\eta(Pop)$, $C$, the clustering coefficient, is smaller, and $S$ is larger, both reducing $[\eta(Pop)]^{2}$, with deleterious effects on potential economic growth. (\ref{eq 61}) implies that there also may be an adverse effect on economic growth if $C$ and $S$ are impaired in the formula for $\eta|\{(PS)\}|$ for a society's network of abstractions.

The promotion during \textit{The Enlightenment} of economic, political and philosophical freedom appears to be consistent with the consequences that such freedoms have for economic productivity. Spinoza wrote many years ago that

\begin{quotation}
``Finally, we have shown not only that this freedom [the freedom to say what they think] can be granted without detriment to public peace, to piety, and to the right of the sovereign, but also that it must be granted if these are to be preserved''  \cite[Ch. 20, p 572]{SpinozaTPT}.
\end{quotation}

\subsection{Increases in life expectancy}\label{Subsubsec Test, Life exp}

The seventh  set of data relate to the observation of increasing life expectancies in the world since 1840. Suppose better public health ideas began to circulate in the 1800s. Individuals evaluate life style choices affecting health---what to eat, whether to exercise, and what other individual lifestyle choices to make---that can affect an individual's longevity. The data suggests that there is no extra $\eta(Pop)$ factor as there is in the case of economic productivity, likely because the evaluation of lifestyle choices is largely for oneself. Does life expectancy increase at the \textit{average} individual problem solving rate? 

The data set out in Table \ref{Table 6} is excerpted from \cite{OeppenSupp} which contains material supplementary to an article by Jim Oeppen and James W. Vaupel \cite{Oeppen2002} about life expectancies. I chose data from \cite{OeppenSupp} for people of the same sex and country at different times. L.E. in Table \ref{Table 6} represents life expectancy for the year to the adjacent left of the L.E. column. The last column sets out the percentage by which the rate of growth per decade in life expectancy exceeds 3.41\% per decade. 

\begin{table}[ht]
\begin{center}
\footnotesize
	\begin{tabular}{|c|c|c|c|c|c|c|c|}\hline
	Country & M-F&$t_{1}$&L.E.&$t_{2}$&L.E.&Rate/dec&cf. 3.41\%\\ \hline
Norway & F & 1841&47.9&1970&77.32&3.71\%&+8.85\%\\ \hline
Norway & M & 1841&44.5&1960&71.39&3.97\%&+16.48\%\\ \hline
New Zealand & M & 1876&51.99&1944&66.58&3.63\%&+6.672\%\\ \hline
Denmark & M & 1840&43.11&1919&56.69&3.46\%&+1.652\%\\ \hline
	\end{tabular}
		\caption{Increases in human longevity, from \cite{OeppenSupp}}  \label{Table 6}
		\end{center}
\end{table}

\normalsize

From Table \ref{Table 6}, the rates of increase in longevity for the periods considered are close to the 3.41\% that appears to be society's long-term average general problem solving rate, but higher as set out in the column in Table \ref{Table 6} headed `cf. 3.41\%.' The discrepancies in rates may be due to the function $h$, the infrastructure factor mentioned in (\ref{eq 28}). The discrepancy may arise because  measurement over a period of about 100 years as in Table \ref{Table 6} has more uncertainty than measurement over a longer period, recalling the error analysis set out in Tables \ref{Table 1-1} and \ref{Table 1-1Lower}.

In connection with their study, Oeppen and James W. Vaupel write, 

\begin{quote}
	``The linear climb of record life expectancy suggests that reductions in mortality should not be seen as a disconnected sequence of unrepeatable revolutions but rather as a regular stream of continuing progress \cite{Oeppen2002}.''
\end{quote}

The `regular stream of continuing progress' is, likely, largely due to the increase in society's average general collective problem solving rate, equal to the increase in the average general individual problem solving rate. These results are consistent with, but not proof of, \textit{The General Collective Problem Solving Capacity Hypothesis}. The rate of increase in life expectancy is consistent with (\ref{eq 25}).

\section{Conclusion}\label{sec End}

Together, the seven sets of data set out above, and particularly the calculations relating to the average innate, basal, or bi-nodal rate and glottochronology, support the concepts and equations described in this article.

This article likely raises more questions than it answers, including the problem of discovering other problems to which these ideas might be applied. It seems to me that a particularly promising area of inquiry is economics because of the availability of economic statistics. It may be that some of these ideas may also be helpful in epidemiology, and in educational issues.

\end{document}